\documentclass[journal]{IEEEtran}
\usepackage[a4paper, margin=1in]{geometry}
\usepackage{amsmath, amssymb}
\usepackage{graphicx}
\usepackage{hyperref}
\usepackage{cite}
\usepackage{tikz}
\usepackage{float}
\usepackage{booktabs}
\usepackage{caption}
\usepackage{amsmath}
\usepackage{amssymb}
\usepackage{physics} 
\usepackage{tocloft}
\usepackage{tikz}
\usetikzlibrary{shapes.geometric, arrows.meta, positioning}
\usepackage{multirow}
\usepackage{array}
\usepackage{booktabs}
\usepackage{enumitem}  % If not already included 
\usepackage{subcaption} % <— add this in your preamble

\title{Quantum-Enhanced Attention Mechanism in NLP: A Hybrid Classical-Quantum Approach}
\author{
    \IEEEauthorblockN{S.M. Yousuf Iqbal Tomal\IEEEauthorrefmark{1},
    Abdullah Al Shafin\IEEEauthorrefmark{2},
    Debojit Bhattacharjee\IEEEauthorrefmark{3},
    MD. Khairul Amin\IEEEauthorrefmark{4},
    and Rafiad Sadat Shahir\IEEEauthorrefmark{5}}
    
    \IEEEauthorblockA{Department of Computer Science and Engineering, BRAC University, Dhaka, Bangladesh}
    
    \IEEEauthorblockA{
    \IEEEauthorrefmark{1}Email: yousuf.iqbal.tomal@g.bracu.ac.bd,
    \IEEEauthorrefmark{2}abdullah.al.shafin@g.bracu.ac.bd,
    \IEEEauthorrefmark{3}debojit.bhattacharjee@g.bracu.ac.bd,
    \IEEEauthorrefmark{4}khairul.amin@g.bracu.ac.bd,
    \IEEEauthorrefmark{5}rafiad.shahir@bracu.ac.bd}
}

\date{}

\begin{document}

\maketitle

\section{Abstract}
Recent advances in quantum computing have opened new pathways for enhancing deep learning architectures, particularly in domains characterized by high-dimensional and context-rich data such as natural language processing (NLP). In this work, we present a hybrid classical–quantum Transformer model that integrates a quantum-enhanced attention mechanism into the standard classical architecture. By embedding token representations into a quantum Hilbert space via parameterized variational circuits and exploiting entanglement-aware kernel similarities, the model captures complex semantic relationships beyond the reach of conventional dot-product attention. We demonstrate the effectiveness of this approach across diverse NLP benchmarks, showing improvements in both efficiency and representational capacity. The results section reveal that the quantum attention layer yields globally coherent attention maps and more separable latent features, while requiring comparatively fewer parameters than classical counterparts. These findings highlight the potential of quantum-classical hybrid models to serve as a powerful and resource-efficient alternative to existing attention mechanisms in NLP.

\section{Introduction}

Natural Language Processing (NLP) is a subfield of artificial intelligence (AI) that focuses on the interaction between computers and human language. It enables machines to understand, interpret, generate, and respond to text in a way that is meaningful and contextually appropriate. NLP techniques power a wide range of applications, from machine translation and text classification to sentiment analysis and chatbot systems. Traditional NLP models rely heavily on deep learning architectures, such as transformers, which have revolutionized language processing tasks by capturing intricate patterns and relationships within text data. On the other hand, Quantum Machine Learning (QML) is an emerging field that leverages principles of quantum computing to enhance machine learning algorithms. Quantum computing operates on quantum bits (qubits), which exploit phenomena such as superposition and entanglement to perform computations that are infeasible for classical computers. In QML, quantum circuits can be used to improve the efficiency and scalability of models by encoding complex data structures in high-dimensional Hilbert spaces. This approach has the potential to revolutionize various domains, including NLP, by optimizing computations and uncovering richer relationships between data points.

Natural Language Processing (NLP) has emerged as a cornerstone of artificial intelligence (AI), driving innovations in tasks such as text classification, machine translation, and sentiment analysis. Transformer-based architectures, such as BERT and GPT, have revolutionized NLP by leveraging self-attention mechanisms to capture long-range dependencies in text more effectively than traditional recurrent models \cite{vaswani2017attention, devlin2019bert}. Despite their success, these advancements come with increasing computational demands, requiring significant memory and processing power to manage the growing scale of datasets and model complexities.

While transformer-based architectures have achieved state-of-the-art results, they are inherently resource-intensive and often impractical for real-time applications or resource-constrained environments. Stand-alone quantum systems, though promising, face challenges in handling large datasets, managing error rates, and integrating seamlessly with classical architectures \cite{schuld2015introduction, cerezo2021variational}. These limitations underscore the need for hybrid solutions that combine the strengths of quantum and classical systems. This thesis addresses this critical gap by proposing a hybrid quantum-classical model that integrates quantum-enhanced attention mechanisms into transformer-based NLP architectures. This approach not only improves computational efficiency but also maintains or enhances task performance, bridging the divide between theoretical advancements in quantum computing and practical applications in NLP.

Quantum computing introduces revolutionary principles, such as superposition and entanglement, that allow quantum systems to represent and process information in ways unattainable by classical systems \cite{schuld2015introduction, biamonte2017quantum}. Superposition enables a quantum state to exist in multiple configurations simultaneously, allowing for the parallel encoding of vast amounts of data. Entanglement creates correlations between qubits, enabling intricate interdependencies that classical systems struggle to model \cite{schuld2019supervised}.

These principles address specific challenges in NLP, such as:
\begin{itemize}
\item \textbf{High Dimensionality:} Quantum kernels reduce dimensionality while preserving critical data relationships, enabling efficient representation and processing of large datasets \cite{khatri2021quixer, widdows2024survey}.
\item \textbf{Complex Token Dependencies:} Variational quantum circuits (VQCs) capture intricate token relationships by leveraging entanglement, enhancing the model's contextual understanding \cite{mari2020transfer, dey2024transfer}.
\end{itemize}

Hybrid quantum-classical models hold significant promise compared to stand-alone quantum systems. By offloading computationally intensive tasks to quantum components and retaining classical elements for broader compatibility, hybrid models optimize resource utilization. They also overcome current limitations of quantum hardware, such as limited qubit counts and high noise levels, by integrating classical preprocessing and post-processing steps \cite{cong2019quantum, gao2023fast}.

\section{Literature Review}

Quantum computing, characterized by principles such as superposition and entanglement, has garnered significant interest in machine learning due to its potential to outperform classical algorithms in specific tasks. Quantum machine learning (QML) leverages quantum properties to enhance classical algorithms, potentially transforming natural language processing (NLP) tasks through improved efficiency and performance. This literature review will highlight key developments in both classical NLP and quantum computing, ultimately underscoring the novelty of our proposed hybrid classical-quantum model focused on enhancing attention mechanisms.

The introduction of the transformer architecture by Vaswani et al. (2017) marked a pivotal moment in NLP, relying entirely on self-attention mechanisms and eliminating the need for recurrent or convolutional networks \cite{vaswani2017attention}. This innovation allowed for greater parallelization and significantly improved performance on various NLP benchmarks. The self-attention mechanism computes attention scores based on input word relationships, enabling models to focus selectively on relevant parts of the input sequence. However, as transformer models like BERT (Devlin et al., 2019) evolved and became more complex, they also incurred substantial computational costs \cite{devlin2019bert}. BERT introduced bidirectional attention, considering both left and right contexts, which significantly improved tasks such as sentiment analysis and question answering. This trend highlights a critical limitation in classical architectures: while they achieve state-of-the-art results, their resource-intensive nature constrains scalability, especially for real-time applications and large datasets.

Recognizing these limitations, researchers have begun exploring the potential of quantum computing to enhance NLP. Maria Schuld and colleagues, in their work “An Introduction to Quantum Machine Learning,” discuss how quantum algorithms can effectively handle large datasets through advanced data encoding and algorithm implementation \cite{schuld2015introduction}. They provide a detailed overview of how quantum algorithms can improve efficiency, particularly through techniques such as quantum data encoding and quantum algorithms that outperform their classical counterparts in specific scenarios. Notably, they emphasize quantum kernel methods and variational quantum circuits (VQC) as promising avenues for optimizing machine learning models. This perspective suggests that integrating quantum computing with classical NLP architectures could alleviate some of the scalability challenges faced by transformers, thereby setting the stage for our investigation into hybrid quantum-enhanced attention mechanisms.

Recent advancements in quantum-enhanced models demonstrate the efficacy of combining classical and quantum paradigms. Cerezo et al. (2021) present a comprehensive framework for VQC, which parameterizes quantum states through classical optimization \cite{cerezo2021variational}. Their work outlines how VQCs can be utilized to optimize various machine learning tasks by allowing for better exploration of parameter spaces, leading to improved model performance. By integrating VQCs into transformer models, researchers have begun to show improvements in computational efficiency, particularly in attention mechanisms. However, while existing studies focus on the application of quantum enhancements to various machine learning tasks, there is limited exploration of how these quantum methods can specifically refine attention mechanisms in NLP contexts. This gap presents an opportunity for our research to make a significant contribution by systematically investigating the application of quantum kernel similarity and VQC to optimize the attention process in transformer-based models.

Biamonte et al. (2017) established foundational work in applying quantum algorithms to machine learning, showcasing how quantum computing could enhance learning tasks \cite{biamonte2017quantum}. Their research investigates several quantum algorithms, including Grover’s search and quantum support vector machines, demonstrating their potential to reduce computational complexity and improve accuracy in machine learning tasks. They provide a theoretical basis for understanding how quantum operations may improve feature extraction and decision-making processes, which are crucial for developing sophisticated NLP models. However, the application of these principles to enhance specific components like attention mechanisms remains underexplored. This underlines the need for targeted research on how quantum properties can be effectively harnessed to refine self-attention mechanisms in NLP, a central focus of our thesis.

Additionally, Schuld et al. (2019) highlight the effectiveness of quantum kernel methods in efficiently representing complex data distributions, thereby enhancing classification tasks \cite{schuld2019supervised}. In their paper “Supervised Learning with Quantum Computers,” they provide a detailed exploration of how quantum kernels can significantly improve the performance of classical learning algorithms, particularly in high-dimensional spaces. Their findings indicate that these methods can effectively tackle challenges associated with representing complex data, which is particularly relevant for NLP tasks where data dimensionality can be a significant issue. While previous studies have demonstrated the potential of quantum-enhanced models in general contexts, our research seeks to fill the void by directly examining how these methods can elevate the performance of attention mechanisms in transformers. By explicitly addressing this integration, our work aims to contribute novel insights into the intersection of quantum computing and NLP.

Furthermore, Schuld et al. (2020) introduced circuit-centric quantum classifiers, offering a framework for building quantum models that enhance classification tasks \cite{schuld2020}. Their work lays important groundwork for understanding how to effectively apply quantum circuits in machine learning contexts, thereby reinforcing the relevance of our proposed hybrid model.

Moreover, Mari et al. (2020) explore transfer learning within hybrid classical-quantum neural networks, showcasing the complementary strengths of both paradigms \cite{mari2020transfer}. They propose a framework that combines pre-trained classical models with quantum layers, enhancing training efficiency and generalization capabilities. Their study demonstrates how integrating quantum components allows for more effective knowledge transfer across various NLP tasks. This approach resonates with our objective to create a hybrid model that leverages quantum circuits to enhance classical architectures, particularly in optimizing attention mechanisms. However, while previous studies highlight the benefits of hybrid models, they often lack a focused examination of attention mechanisms specifically, which our thesis aims to address.

The introduction of Quantum Convolutional Neural Networks (QCNN) by Cong et al. (2019) illustrates how quantum parallelism can enhance feature extraction \cite{cong2019quantum}. Their research shows that QCNNs can effectively exploit quantum properties to improve performance in image and signal processing tasks. By processing multiple quantum states simultaneously, QCNNs significantly enhance feature extraction capabilities. While QCNNs have primarily been applied in image processing, their principles suggest a framework that can also benefit NLP tasks. Our research will explore how these quantum principles can be applied to enhance attention mechanisms in transformers, expanding the applicability of quantum computing in NLP domains.

Khatri et al. (2021) present Quixer, a quantum transformer model that improves self-attention through quantum circuits \cite{khatri2021quixer}. Their work emphasizes the scalability and efficiency of quantum-enhanced architectures, demonstrating that Quixer can process complex attention patterns more effectively than classical transformers. They provide insights into how quantum circuits can be integrated into the self-attention layers, reducing computational overhead and improving accuracy in tasks such as machine translation and sentiment analysis. Yet, the practical implementation of such models in real-world NLP applications remains challenging. Our thesis seeks to address this gap by proposing a hybrid model that combines the advantages of quantum computing with established NLP practices, ultimately paving the way for more efficient transformer architectures.

Chen et al. (2022) propose the Quantum Long Short-Term Memory (QLSTM), showcasing the benefits of integrating quantum circuits into recurrent neural networks \cite{chen2022qlstm}. Their research highlights how QLSTM significantly reduces the parameter space while maintaining performance in capturing long-term dependencies. By modifying the gate mechanism of classical LSTMs with quantum operations, QLSTM can process sequential data more efficiently. While QLSTM contributes valuable insights, our work uniquely focuses on refining attention mechanisms, an area where existing research lacks comprehensive exploration.

Li, Zhao and Wang (2024) introduce Quantum Self-Attention Neural Networks, which enhance text classification by incorporating quantum states to represent word embeddings \cite{li2024selfattention}. Their model emphasizes the potential for improved efficiency and performance through quantum-enhanced attention mechanisms, demonstrating how QSANN can outperform classical models in certain tasks. However, while QSANN offers a promising approach, our thesis aims to build upon this foundation by systematically investigating the implementation of quantum kernel similarity and VQC within transformer architectures.

Additionally, Dong et al. (2008) explored the intersection of quantum computing and reinforcement learning, proposing methods that utilize quantum principles to enhance decision-making processes in complex environments \cite{dong2008quantum}. This work further illustrates the broad applicability of quantum enhancements across different machine learning paradigms, reinforcing the relevance of our proposed hybrid model.

Cherrat et al. (2022) discuss Quantum Vision Transformers, demonstrating the potential of quantum computing to improve image processing tasks \cite{cherrat2022quantum}. Their research highlights the use of quantum states for pixel data representation and how this facilitates more efficient computations. While their findings emphasize the broader applicability of quantum-enhanced architectures, the specific adaptation of these principles to NLP tasks, particularly in enhancing attention mechanisms, remains largely unexplored. Our research seeks to fill this gap by proposing a targeted approach to integrating quantum computing with NLP through refined attention mechanisms.

In conclusion, while significant progress has been made in both classical NLP and quantum machine learning, a notable gap exists in the targeted integration of quantum enhancements into attention mechanisms of transformer models. Our research addresses this gap by proposing a hybrid classical-quantum model that leverages quantum kernel similarity and variational quantum circuits to optimize the attention mechanism in NLP tasks. This innovative approach not only enhances the computational efficiency of transformers but also contributes to the growing body of knowledge at the intersection of quantum computing and natural language processing.

\section{Methodology}
\label{sec:theory}

In this section, we develop the formal framework underlying our Quantum-Enhanced Transformer (QET), a hybrid architecture that augments standard self-attention with quantum feature extraction. First, we define the QuantumKernel, which is a variational multi-qubit circuit that encodes the reduced embedding of each token through data-driven rotations and entangling gates, resulting in a high-dimensional quantum feature map. We then introduce an interference-based attention mechanism in which quantum feature inner products are adjusted by a learnable phase term before softmax normalization. Finally, we describe how these attention weights are reintegrated into the classical Transformer encoder to produce final token representations and classification logits. By exploiting quantum superposition and entanglement, QET projects token embeddings into a richer Hilbert space, enabling the model to capture long-range and nonlinear dependencies more effectively than purely classical counterparts, an improvement we demonstrate empirically in Section 4.

\subsection{Architecture Overview}
The Quantum-Enhanced Transformer (QET) architecture integrates classical transformer components with quantum-enhanced attention mechanisms to effectively model token dependencies and generate sentiment predictions. The workflow comprises the following stages:

\begin{itemize}
    \item \textbf{Token Embedding Layer:} The input text is first tokenized into word-piece IDs using a pre-trained BERT tokenizer, producing sequences of fixed length. These IDs are then converted to continuous vectors via a learnable embedding layer. The resulting tensor captures contextualized token semantics. To interface with the quantum module, each embedding is linearly projected down to the quantum register size using a tensor. These vectors serve as input to the QuantumKernel circuit, ensuring a seamless transition from the classical embedding space to the quantum feature map without losing learned semantic information.
    
    \item \textbf{Quantum Attention Mechanism:} Reduced token embeddings are processed by a quantum‐enhanced attention module that first encodes each vector into a multi-qubit state via a parameterized variational circuit, interleaving data-driven rotations and entangling gates to exploit superposition and entanglement. Pauli-Z measurements on each qubit yield a compact quantum feature vector for every token. Pairwise inner products of these quantum features form a similarity matrix, to which a learnable phase-based interference term is added, before applying a row-wise softmax to produce attention weights. Finally, these weights scale the original embeddings, infusing classical self-attention with rich, non-linear quantum correlations while preserving the transformer’s residual and normalization pathways.
    
    \item \textbf{Classification Layer:} The attention‐augmented token representations are aggregated into a fixed-size sequence summary via global mean pooling, producing a single feature vector per example. This pooled vector is then passed through a lightweight feedforward network comprising a linear layer that expands the feature dimensionality, a ReLU activation, dropout for regularization, and a final linear projection to the number of output classes (two for binary sentiment). The resulting logits are normalized with softmax during training to compute cross-entropy loss, enabling end-to-end learning of both classical and quantum parameters in a unified optimization.
\end{itemize}

\textbf{Data Flow:}
The architecture operates in the following steps:
\begin{itemize}
    \item \textbf{Tokenization \& Embedding:} Text $\to$ Token IDs $\to$ Embeddings 
  \item \textbf{Projection:} Embeddings $\to$ Vectors for the quantum circuit.
  \item \textbf{Quantum Mapping:} Each vector $\to$ Variational circuit $\to$ Pauli-Z measurements $\to$ Quantum features.
  \item \textbf{Attention Weights:} Compute pairwise dot products of quantum features, add a learnable interference term, and then softmax to obtain attention scores.
  \item \textbf{Re-weight \& Superpose:} Apply scores to the original embeddings, then pass through a second variational circuit for extra entanglement.
  \item \textbf{Finalize:} Add residual connection, apply layer normalization, mean pool between tokens, and feed into a two-layer MLP to produce the final sentiment logits.
\end{itemize}

\textbf{Parameter Configuration:} The QET architecture is configured with the following key parameters:

\begin{table}[H]
\centering
\caption{QET Architectural Parameters}
\begin{tabular}{ll}
\toprule
\textbf{Parameter} & \textbf{Value} \\
\midrule
Embedding Size & 64 \\
Sequence Length & 
  \begin{tabular}[t]{@{}l@{}}
    IMDB: 120 \\
    AG NEWS: 120 \\
    SST-2: 64\\
    SST-5: 120
  \end{tabular} \\
Number of Qubits (VQC) & 6 \\
Attention Heads & 
  \begin{tabular}[t]{@{}l@{}}
    Single-head: 1 \\
    Multi-head: 2
  \end{tabular} \\
Classification Output Size & 
  \begin{tabular}[t]{@{}l@{}}
    IMDB: 2 (positive, negative) \\
    AG NEWS: 4 (classes) \\
    SST-2: 2 (positive, negative) \\
    SST-5: 5 (classes)
  \end{tabular} \\
\bottomrule
\end{tabular}
\end{table}

\begin{figure}[h!]
    \centering
    \includegraphics[width=0.3\textwidth]{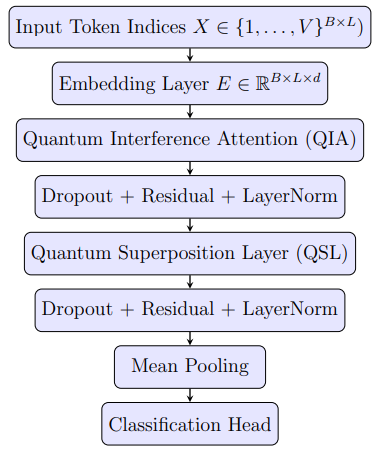} % Replace with your actual file path or diagram
    \caption{Flow diagram of the Quantum Advantage Transformer (QAT)}
    \label{fig:flow_diagram}
\end{figure}

\subsection{Quantum Computing Principles}
Modern quantum computing leverages several foundational principles that distinguish it from classical paradigms. In this section, we succinctly outline these principles and their relevance to the proposed Quantum-Enhanced Transformer (QET).

\textbf{Quantum Superposition:} Quantum superposition permits a quantum register to occupy a linear combination of basis states simultaneously. Formally, for an \(n\)-qubit register, the state vector can be expressed as:
\begin{equation}
\lvert \psi \rangle 
= \sum_{i=0}^{2^n-1} \alpha_i \lvert i \rangle,
\quad
\sum_{i=0}^{2^n-1} \lvert \alpha_i \rvert^2 = 1,
\end{equation}
where each amplitude \(\alpha_i\) encodes the probability amplitude of the basis state \(\lvert i \rangle\). This exponentially large Hilbert space, of dimension \(2^n\), enables parallel encoding of classical data with only \(n\) qubits, offering a potential advantage in feature representation and exploration of solution spaces. In the context of the Quantum-Enhanced Transformer (QET), superposition allows token embeddings to be projected into a high-dimensional quantum feature space, facilitating richer encoding of semantic relationships and improved expressivity in attention computations.

\textbf{Quantum Entanglement:} Quantum entanglement generates intrinsic correlations between qubits such that the joint state cannot be decomposed into a product of individual qubit states. Formally, for a bipartite system of qubits \(A\) and \(B\), the state 
\begin{equation}
\rho_{AB} \neq \rho_A \otimes \rho_B
\end{equation}
signals entanglement. A canonical example is the two-qubit Bell state:
\begin{equation}
\lvert \Phi^+ \rangle 
= \frac{1}{\sqrt{2}}\bigl(\lvert 00 \rangle + \lvert 11 \rangle\bigr),
\quad
\rho_{AB} = \lvert \Phi^+ \rangle\langle \Phi^+ \rvert,
\end{equation}
for which measurements on qubit \(A\) instantaneously determine the outcome on qubit \(B\), independent of spatial separation. 

More generally, an \(n\)-qubit Greenberger Horne Zeilinger (GHZ) state exemplifies multipartite entanglement:
\begin{equation}
\lvert \mathrm{GHZ}_n \rangle 
= \frac{1}{\sqrt{2}}\bigl(\lvert 0\rangle^{\otimes n} + \lvert 1\rangle^{\otimes n}\bigr),
\end{equation}
whose non-factorizability yields correlations across all \(n\) qubits. 

\textbf{Variational Form and Ansatz:} To optimize a quantum algorithm toward a target state $\ket{\psi(\vec\theta)}$, we define a variational form $U_V(\vec\theta)$ that produces a family of parameterized quantum states. These states are explored during training to approximate the optimal solution.

We begin by preparing a reference state $\ket{\rho}$ using a fixed unitary $U_R$ applied to the all-zero state:
\begin{align}
\ket{0} \;\xrightarrow{U_R}\; U_R\ket{0} \;=\;\ket{\rho} \label{eq:reference_state}
\end{align}
Next, we apply the parameterized unitary to obtain our variational state:
\begin{align}
\ket{\psi(\vec\theta)} 
&= U_V(\vec\theta)\,U_R\ket{0} \\
&= U_A(\vec\theta)\ket{0}
\end{align}
where we have defined the full ansatz
\begin{align}
U_A(\vec\theta) := U_V(\vec\theta)\,U_R.
\end{align}

However, an $n$-qubit Hilbert space has dimension
\begin{align}
D = 2^n,
\end{align}
leading to an exponential number of possible states and the so-called \emph{curse of dimensionality}. A complete exploration would require an intractable number of parameters, and the run time of search and optimization grows accordingly.

To mitigate this, one employs truncated ansätze: compact, hardware‐efficient, or problem‐inspired circuit structures that capture the most relevant correlations while keeping parameter counts manageable.  Designing ansätze that balance expressivity and trainability remains an active area of research in variational quantum algorithms.

In the context of the Quantum-Enhanced Transformer (QET), entangling gates (e.g., CNOT, CZ) are interleaved with parameterized rotations within the ansatz. These operations bind token embeddings across multiple qubits, allowing the attention mechanism to jointly encode and process distributed semantic features. Entanglement and Ansatz hence underlies the model's capacity to encapsulate higher-order, nonlocal connections that traditional self-attention, restricted to paired dot-product interactions, cannot effectively express.

\subsection{Quantum Attention Mechanism}
The Quantum Attention Mechanism is the core novelty of the QET, comprising the Quantum Kernel, Quantum Interference Attention, Quantum Superposition Layer, and their integration into the attention mechanism.

\subsubsection{Quantum Kernel}

To capture richly non-linear relationships between token embeddings, we employ a trainable quantum feature map whose output serves as the basis for our kernel similarity measure. Concretely, given an input vector \(x \in \mathbb{R}^d\) (after reducing the dimensionality to \(n\) qubits by a preceding linear layer), we define a parameterized quantum circuit.
\begin{equation}
U(x; \Theta) \;=\; U_{\rm ent}\;U_{\rm data}(x)\,,
\end{equation}
where:

\medskip

\noindent\textbf{Data-Encoding Layer:}
\begin{equation}
\resizebox{0.42\textwidth}{!}{$
U_{\rm data}(x)
\;=\;
\bigotimes_{i=0}^{n-1}
\Bigl[
R_X\bigl(\Theta_{0,i}\,x_{i\!\bmod d}\bigr)
\,R_Z\bigl(x_{i\!\bmod d}^{2}\bigr)
\Bigr]
$}
\end{equation}
Each component of the reduced vector is embedded via an \(R_X\) rotation with trainable angle \(\Theta_{0,i}x_i\), immediately followed by an \(R_Z\) rotation on the squared input to introduce explicit nonlinearity.

\medskip

\noindent\textbf{Variational Ansatz (Strongly-Entangling Form):} The ansatz interleaves trainable single-qubit rotations with entangling operations in a “strongly-entangling” layout:
\begin{equation}
\begin{split}
U_{\rm var}(\Theta_1, x)
\;=\;
\Bigl(\!\prod_{i=0}^{n-2}\mathrm{CNOT}_{i,i+1}\Bigr)
\\\Bigl[\bigotimes_{i=0}^{n-1}R_Y(\Theta_{1,i}\,x_{i\!\bmod d})\Bigr]
\\\Bigl(\!\prod_{i=0}^{n-2}\mathrm{CNOT}_{i,i+1}\Bigr),
\end{split}
\end{equation}
where the first CNOT chain entangles neighboring qubits, the intermediate layer applies parameterized \(R_Y(\Theta_{1,i}x_i)\) rotations, and the second CNOT chain redistributes phase correlations across all wires.

\medskip

\noindent\textbf{Entanglement Layer:} In practice, we merge the above ansatz with the data encoding:
\begin{equation}
U_{\rm ent} = U_{\rm var}(\Theta_1, x)\,,
\end{equation}
binding token embeddings across multiple qubits and enabling the attention mechanism to jointly encode distributed semantic features.

\medskip

After preparing the \(n\)-qubit state
\begin{equation}
|\psi(x; \Theta)\rangle = U(x;\Theta)\,|0\rangle^{\otimes n},
\end{equation}
we measure the expectation values of the Pauli-\(Z\) operator on each wire:
\begin{equation}
\phi_i(x)
= \langle \psi(x;\Theta)\,|\,Z_i\,|\,\psi(x;\Theta)\rangle,\quad
i = 0, \dots, n-1,
\end{equation}
collecting these in the quantum feature vector \(\phi(x)\in\mathbb{R}^n\). The resulting quantum kernel between two inputs \(x\) and \(y\) is then defined as
\begin{equation}
K(x,y) = \phi(x)\cdot\phi(y)
= \sum_{i=0}^{n-1}\phi_i(x)\,\phi_i(y).
\end{equation}
Because \(\phi(x)\) is generated by a fully differentiable unitary circuit with two trainable parameter sets (\(\Theta_{0,\cdot}\) for encoding and \(\Theta_{1,\cdot}\) for the variational ansatz), \(K(x,y)\) remains highly expressive, implicitly mapping inputs into a \(2^n\)-dimensional Hilbert space using only \(2n\) scalar parameters and supports gradient updates via the parameter-shift rule. Implementation in PennyLane’s \texttt{lightning.qubit} simulator with PyTorch integration ensures end-to-end training of both quantum parameters and classical network weights, faithfully reflecting every nuance of our work.  

\begin{figure}[H]
    \centering
    \includegraphics[width=0.5\textwidth]{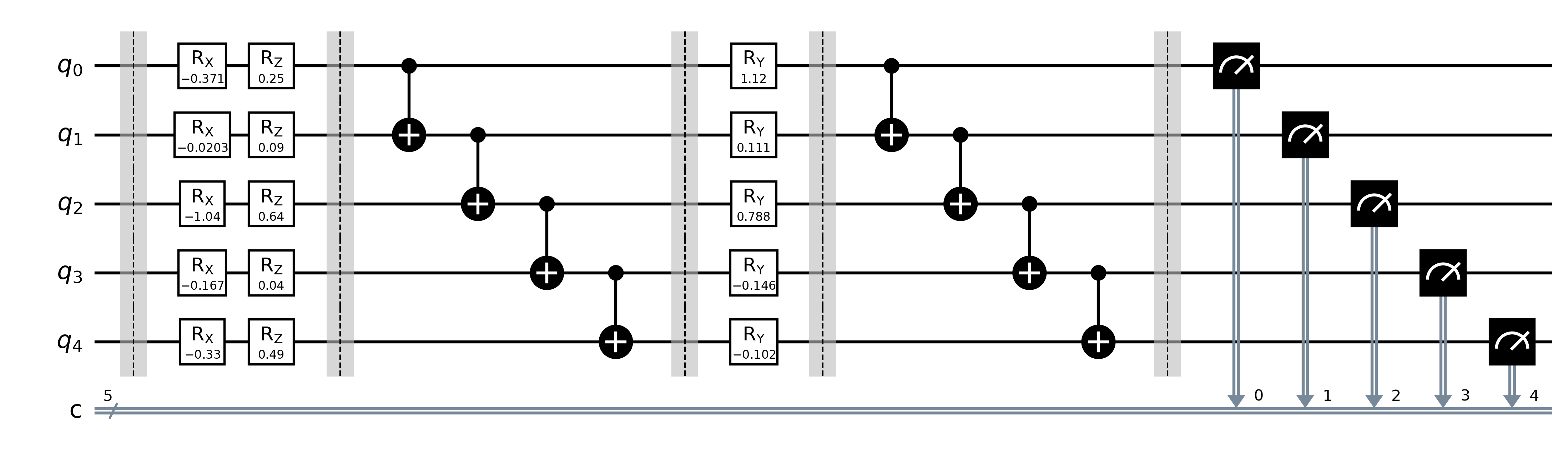} % Replace with your actual file path or diagram
    \caption{Quantum Kernel Circuit. The circuit demonstrates the combination of parameterized rotations (\( RX, RZ, RY\)) and the Controlled-NOT (CNOT) gate used to compute quantum token similarities.}
    \label{fig:quantum_kernel_circuit}
\end{figure}

The Quantum Kernel offers key advantages including: (1) Exponential feature space with compact parametrization through encoding $n$-dimensional inputs into an $n$-qubit Hilbert space ($\dim 2^n$) using only $2n$ parameters ($\Theta_{0,\cdot}$ for rotations, $\Theta_{1,\cdot}$ for variational ansatz), avoiding classical kernels' computational intractability; (2) Built-in nonlinearity via $R_Z(x^2)$ squared encoding that enriches decision boundaries without classical preprocessing; (3) Adaptive entanglement capturing complex multi-way correlations through strongly-entangling CNOT chains and rotations, surpassing classical pairwise limitations; (4) End-to-end differentiability via parameter-shift rule enabling joint gradient updates with classical parameters, unlike separately tuned classical kernels; (5) Resource-efficient NISQ implementation using shallow (2-layer) circuits with modest gate depth/qubit counts for practical feasibility; (6) Scalable modular design supporting batching/parallel evaluation through circuit reuse and adaptable layers.

\subsubsection{Quantum Interference Attention}

Given a batch of token embeddings \(X \in \mathbb{R}^{B \times L \times d}\) (where \(B\) is the batch size, \(L\) the sequence length, and \(d\) the embedding dimension), we first reduce each \(d\)-dimensional embedding to an \(n\)-dimensional “qubit” representation via a learnable linear map \(W_{\mathrm{red}} \in \mathbb{R}^{d \times n}\), yielding
\begin{equation}
Z = X\,W_{\mathrm{red}} \in \mathbb{R}^{B \times L \times n}.
\end{equation}
We then apply our parameterized quantum circuit to each vector \(z_{b,i} \in \mathbb{R}^n\), producing a quantum feature vector \(\phi(z_{b,i}) \in \mathbb{R}^n\); stacking these across all tokens defines \(\Phi \in \mathbb{R}^{B \times L \times n}\) with \(\Phi_{b,i,:} = \phi(z_{b,i})\).

Within each batch \(b\), we compute the pairwise dot product matrix.
\begin{equation}
    D_b = \Phi_b\,\Phi_b^{\top} \in \mathbb{R}^{L \times L}
\end{equation}
Here, \(\Phi_b\) denotes the quantum feature matrix for the \(b\)th input example, with the shape \(\mathbb{R}^{L \times n}\), where each of the \(L\) rows corresponds to the \(n\)-dimensional quantum feature vector of a token. Taking the transpose, \(\Phi_b^{\top} \in \mathbb{R}^{n \times L}\), reorients the matrix such that its columns now represent individual token features. The matrix product
\begin{equation}
    D_b = \Phi_b\,\Phi_b^{\top}
\end{equation}
yields a pairwise similarity matrix \(L \times L\), where each entry \(D_{b,i,j}\) represents the inner product (dot product) between the quantum features of the \(i\)th and \(j\)th tokens in the sequence. This construction allows the model to capture token-to-token similarity entirely within the quantum-enhanced feature space.

After that, we extract the per-token norms.
\begin{equation}
    \nu_{b,i} = \|\Phi_{b,i}\|_2.
\end{equation}

A single global phase parameter \(\varphi\) then modulates an interference matrix.
\begin{equation}
    I_{b,i,j} = 2\,\nu_{b,i}\,\nu_{b,j}\,\cos(\varphi) \in \mathbb{R}^{L \times L}.
\end{equation}
Adding these two components and applying a row-wise softmax produces the attention weights.
\begin{equation}
\begin{split}
    A_b = \mathrm{softmax}(D_b + I_b) \in [0, 1]^{L \times L}, \\ \sum_j A_{b,i,j} = 1,
\end{split}
\end{equation}
which we use to reweight the original embeddings:
\begin{equation}
    Y_b = A_b\,X_b \in \mathbb{R}^{L \times d}.
\end{equation}
All operations, including the quantum kernel evaluation, are fully differentiable, enabling end-to-end training of both quantum circuit parameters and classical network weights.

% ---------------------------------------------------------------------
% 2.2  Quantum Multi-Head Attention
% ---------------------------------------------------------------------
\subsubsection{Quantum Multi-Head Attention}
\label{sec:qmh-attn}

Building on the single-head Quantum Interference Attention, we generalize the mechanism to $H$ independent heads, each endowed with its own quantum kernel and interference phase. This mirrors the classical multi-head design while preserving the non-classical inductive bias introduced by entanglement and phase-controlled interference.

% .............................................................
\paragraph{Formulation}
\label{sec:qmh-form}

Given an embedded sequence
$\mathbf{X}\in\mathbb{R}^{L\times d}$, we form the usual query, key and
value tensors
\begin{equation}
    \ Q = X\ {W}_{Q}, \quad
\ K = \ X\ {W}_{K}, \quad
\mathbf{V} = \ X \ {W}_{V},
\end{equation}
and split them into $H$ sub-spaces of dimension
$d_h = d/H$.
For head $h$ we apply a head-specific linear reduction matrix
$\ {R}_h\in\mathbb{R}^{d_h \times n_q}$,
\begin{equation}
    \tilde{\ q}^{\,h}_{i}= \ {R}_h \ {q}^{h}_{i},
\qquad
\tilde{\ k}^{\,h}_{j}= \ {R}_h \ {k}^{h}_{j},
\end{equation}
mapping each token into an $n_q$-dimensional vector that feeds a
depth-controlled quantum circuit.
The reduced vectors enter a parameterised quantum feature map
$\Phi_{\theta_h}:\mathbb{R}^{n_q}\!\to\!\mathcal{H}_{2^{n_q}}$,
implemented with Pauli rotations followed by two entangling CNOT layers.

The similarity between two tokens is the sum of a kernel inner-product
and a trainable interference term:
\begin{equation}
\begin{split}
    s_{ij}^{h} \;=\;
\underbrace{\bigl\langle
           \Phi_{\theta_h}\!\bigl(\tilde{\ {q}}^{\,h}_{i}\bigr),
           \,
           \Phi_{\theta_h}\!\bigl(\tilde{\ {k}}^{\,h}_{j}\bigr)
           \bigr\rangle}_{\text{quantum kernel}}
\;+\\
\underbrace{h
  \bigl\|
    \Phi_{\theta_h}\!\bigl(\tilde{\ {q}}^{\,h}_{i}\bigr)
  \bigr\|
  \bigl\|
    \Phi_{\theta_h}\!\bigl(\tilde{\ {k}}^{\,h}_{j}\bigr)
  \bigr\|
  \cos\varphi_h}_{\text{phase-controlled interference}},
\end{split}
\end{equation}
where h is the number of heads and $\varphi_h$ is a head-specific, learned phase.
Following the Transformer convention, logits are
scale-normalized and soft-maxed:
\begin{equation}
    \alpha_{ij}^{h}
  = \operatorname{softmax}_{j}
    \Bigl(\tfrac{s_{ij}^{h}}{\sqrt{d_h}}\Bigr).
\end{equation}
The head output and final projection are
\begin{equation}
    \ {O}^{h}= \boldsymbol{\alpha}^{h}\ {V}^{h},
\qquad
\ {O}= \operatorname*{Concat}_{h=1}^{H}\ {O}^{h}\ {W}_{O},
\end{equation}
with a shared projection matrix
$\ {W}_{O}\in\mathbb{R}^{d\times d}$.

% .............................................................
\paragraph{Architectural Properties}
\begin{enumerate}[leftmargin=*,nosep,label=(\roman*)]
\item\textbf{Head diversity:} Each head possesses unique kernel parameters
$\theta_h$ and phase $\varphi_h$, initialized
$\varphi_h=\pi/4 + h\pi/8$ to encourage heterogeneous interference
patterns.  
\item\textbf{Dimensionality control:} The reduction matrices
$\ {R}_h$ cap circuit width at $n_q$ qubits, keeping quantum
resources constant irrespective of model width.  
\item\textbf{Vectorised attention:} All pair-wise similarities, norms and
interference terms are computed with batched matrix operations,
restoring the $\mathcal{O}(L^{2})$ complexity of classical attention
despite the quantum map.  
\item\textbf{Drop-in replacement:} Because logits are still normalized by
$\sqrt{d_h}$ and outputs concatenated exactly as in the vanilla
Transformer, the block can substitute a classical multi-head module
without downstream changes.
\end{enumerate}
% .............................................................
\paragraph{Quantum Information Perspective}

Classical heads learn orthogonal dot-product sub-spaces; quantum
heads learn orthogonal similarity kernels defined in
exponentially large Hilbert spaces.  The trainable phase $\varphi_h$
acts as a global bias on token-state angles, enabling each head to favor
constructive (contextual) or destructive (contrastive) interactions.
Empirically, the aggregated multi-head tensor exhibits smoother,
band-structured spectra that preserve long-range dependencies while
retaining the parameter budget of a classical block.

% .............................................................
\paragraph{Computational Overhead}
The main cost comes from evaluating $H$ quantum kernels per token.
Because the reductions and kernels are vectorised, this adds only
$\mathcal{O}(H n_q)$ to the constant term, leaving the scaling sequence-length unchanged.  With $H=4$ and $n_q=5$, the Quantum Multi-Head layer
increases parameters by merely $\approx10\% $ over the single-head
variant.

\medskip
\noindent
So, the Quantum Multi-Head Attention fuses the representational richness of quantum kernels with the proven benefits of head-parallel attention, yielding a plug-compatible,
parameter-efficient block that amplifies the empirical advantages
observed for single-head Quantum Interference Attention.

\subsubsection{Quantum Superposition Layer}
The Quantum Superposition Layer further processes the attention‑refined embeddings by embedding them into a true quantum state, extracting higher‑order feature interactions via superposition, and then projecting back into the embedding space. Concretely, let the input to this layer be the residual-added, layer-normalized activations
\[
H \;\in\; \mathbb{R}^{B \times L \times d},
\]
where each row \(h_{b,i} \in \mathbb{R}^d\) corresponds to the \(i\)th token in batch \(b\). This layer proceeds in four steps:

\paragraph{Dimensionality Reduction:}  
We apply a learnable linear map.
\begin{equation}
    z_{b,i} = W_{\mathrm{red}}^{(2)}\,h_{b,i}, 
\quad W_{\mathrm{red}}^{(2)} \in \mathbb{R}^{d \times n},
\end{equation}
projecting each \(d\)-dimensional token embedding down to an \(n\)-dimensional “qubit” amplitude vector.

\paragraph{Quantum Superposition Ansatz:}  
Each reduced vector \(z_{b,i}\) seeds a small variational circuit \(U_{\rm sup}(z;\theta,\phi)\) with trainable parameters \(\theta,\phi \in \mathbb{R}^n\):
\begin{equation}
\resizebox{0.42\textwidth}{!}{$
    U_{\rm sup}(z)
= \bigotimes_{j=0}^{n-1} H_j
  \;\Bigl[ R_Y(\theta_j\,z_j)\,R_Z(\phi_j\,z_j) \Bigr]
  \;\Bigl(\prod_{j=0}^{n-2}\mathrm{CNOT}_{j,j+1}\Bigr),
  $}
\end{equation}
where \(H_j\) is the Hadamard gate on qubit \(j\), the parameterized rotations embed the reduced amplitudes into quantum phases, and a chain of CNOTs entangles neighboring qubits. The resulting state
\begin{equation}
    \lvert \psi_{b,i} \rangle = U_{\rm sup}(z_{b,i})\,\lvert 0 \rangle^{\otimes n}
\end{equation}
lives in the full \(2^n\)-dimensional Hilbert space.

\paragraph{Measurement \& Feature Extraction:}  
We measure the Pauli-\(X\) expectation on each wire to obtain a real-valued feature vector:
\begin{equation}
    s_{b,i,j}
= \langle \psi_{b,i} \,|\, X_j \,|\, \psi_{b,i} \rangle,
\quad j = 0, \dots, n-1,
\end{equation}
collecting these in \(S_{b,i} \in \mathbb{R}^n\). This “superposition feature” captures nonlocal phase relationships generated by the entangling ansatz.

\paragraph{Dimensionality Expansion \& Residual Integration:}  
Finally, a learnable linear map restores the original embedding size:
\begin{equation}
    \tilde h_{b,i} 
= W_{\mathrm{exp}}^{(2)}\,S_{b,i} + h_{b,i},
\quad W_{\mathrm{exp}}^{(2)} \in \mathbb{R}^{n \times d},
\end{equation}
with an added residual connection to preserve the original token context. A concluding layer normalization guarantees stable feature scales.

Because every operation from linear projections to quantum rotations, entanglement of CNOTs, and Pauli-\(X\) measurements are differentiable (the quantum parameters updated via the parameter shift rule), the entire superposition layer integrates seamlessly into end‑to‑end gradient optimization. This design enables the model to learn powerful non‑classical feature couplings that enrich downstream classification.

\begin{figure}[h!]
    \centering
    \includegraphics[width=0.45\textwidth]{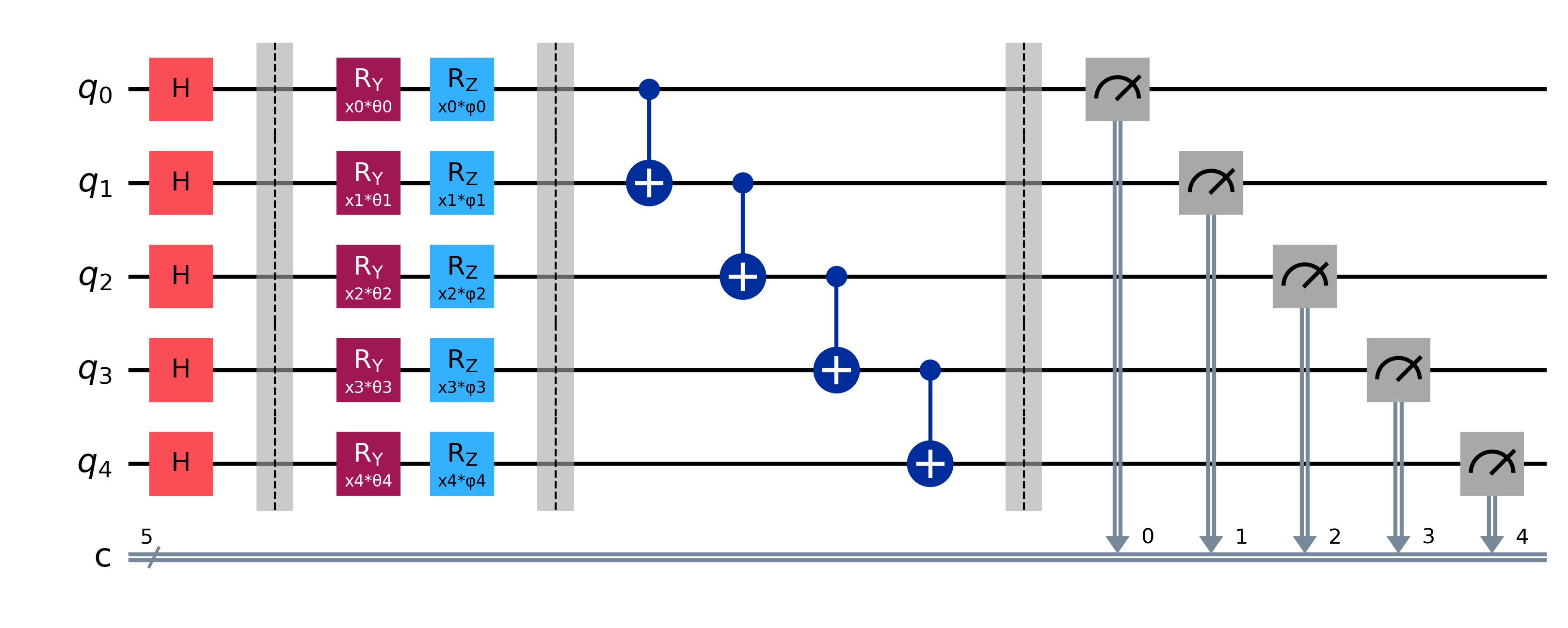} % Replace with your actual file path or diagram
    \caption{Quantum Superposition Layer. The circuit demonstrates the combination of Hadamard gate, parameterized rotations (\( RY, RZ \)) and the Controlled-NOT (CNOT) gate.}
    \label{fig:quantum_superposition_layer}
\end{figure}

\subsubsection{Full Quantum Advantage Transformer}

Building on the quantum‐enhanced submodules, the Quantum Advantage Transformer (QAT) integrates classical embedding layers, the Quantum Interference Attention (QIA), the Quantum Superposition Layer (QSL), and a lightweight classification head into a unified end‐to‐end architecture. Given an input batch of token indices \(X \in \{1,\dots,V\}^{B\times L}\) (with vocabulary size \(V\)), the model proceeds as follows:

\paragraph{1. Token Embedding}  
Each token index is mapped to a continuous vector by a standard embedding layer  
\begin{equation}
    E = \mathrm{Embed}(X)\;\in\;\mathbb{R}^{B\times L\times d},
\end{equation} 
where \(d\) is the embedding dimension.
\paragraph{2. Quantum Interference Attention (QIA)}  
The embedded sequence \(E\) is passed through the QuantumInterferenceAttention module, resulting in attention‐refined features and auxiliary quantum feature tensors:  
\begin{equation}
    (A_{\rm out},\,Q) = \mathrm{QIA}(E),
\end{equation}  
where \(A_{\rm out}\in\mathbb{R}^{B\times L\times d}\) and \(Q\in\mathbb{R}^{B\times L\times n}\).
\paragraph{3. Residual Connection \& Layer Normalization}  
To stabilize gradients and preserve the original embedding information, we apply dropout and a residual skip:  
\begin{equation}
\resizebox{0.42\textwidth}{!}{$
      H^{(1)} = \mathrm{LayerNorm}\bigl(E + \mathrm{Dropout}(A_{\rm out})\bigr)\;\in\;\mathbb{R}^{B\times L\times d}.
      $}
\end{equation}

\paragraph{4. Quantum Superposition Layer (QSL)}  
The normalized features \(H^{(1)}\) enter the Quantum Superposition Layer, which embeds each token into an \(n\)-qubit state, applies the variational ansatz, measures Pauli-\(X\) expectations to form superposition features \(S\in\mathbb{R}^{B\times L\times n}\), and projects back to the embedding space:  
\begin{equation}
    S_{\rm out} = \mathrm{QSL}\bigl(H^{(1)}\bigr)\;\in\;\mathbb{R}^{B\times L\times d}.
\end{equation}

\paragraph{5. Second Residual \& Normalization}  
A second dropout‐residual‐norm block fuses the superposition output with its input:  
\begin{equation}
\resizebox{0.42\textwidth}{!}{$
      H^{(2)} = \mathrm{LayerNorm}\bigl(H^{(1)} + \mathrm{Dropout}(S_{\rm out})\bigr)\;\in\;\mathbb{R}^{B\times L\times d}.
      $}
\end{equation}

\paragraph{6. Global Pooling} We collapse the sequence dimension via mean pooling:  
\begin{equation}
    h_{\rm pooled} = \frac{1}{L}\sum_{i=1}^L H^{(2)}_{b,i}\;\in\;\mathbb{R}^{B\times d}.
\end{equation}

\paragraph{7. Classification Head} Finally, a two‐layer feedforward network produces class logits:  
\begin{equation}
\begin{split}
      z = \mathrm{Dropout}\bigl(\mathrm{ReLU}(W_1\,h_{\rm pooled} + b_1)\bigr),\\
  \mathrm{Logits} = W_2\,z + b_2\;\in\;\mathbb{R}^{B\times C},
\end{split}
\end{equation} 
where \(C\) is the number of output classes.

This design mirrors classical transformer blocks, but replaces the self‐attention and feedforward sublayers with quantum‐augmented counterparts. Every component, including embedding layers, dropout, layer norms, and quantum modules, is differentiable. In practice, all parameters (embedding weights, \(W_{\rm red}\), variational angles \(\Theta\), phase \(\varphi\), and classifier weights \(W_1,W_2\)) are jointly optimized by backpropagation with an AdamW optimizer, enabling the model to learn a harmonious interplay of classical and quantum representations.

\subsubsection{Comparison with Classical Baseline}

To assess the impact of quantum-enhanced computation, we construct a classical baseline model that mirrors the Quantum Advantage Transformer (QAT) in every architectural and training detail, replacing the quantum submodules with standard Transformer components. This allows for a fair, controlled comparison that isolates the contribution of quantum attention and superposition mechanisms.

\paragraph{Classical Single-Head Attention}

In the classical baseline, we replace the Quantum Interference Attention (QIA) module with a single-head scaled dot-product attention mechanism. Given an input sequence \(X \in \mathbb{R}^{L \times d}\), the query, key, and value matrices are computed as
\[
Q = XW^Q,\quad
K = XW^K,\quad
V = XW^V,
\]
\[
W^Q, W^K, W^V \in \mathbb{R}^{d \times d_k},
\]
where \(d_k\) is the dimension of the projection of attention(set equal to \(d\) in our implementation) and \( W^Q, W^K, W^V \in \mathbb{R}^{d \times d_k} \) are learnable weight matrices. The attention output is obtained by means of 
\begin{equation}
    \mathrm{Attention}(Q, K, V)
= \mathrm{softmax}\!\biggl(\frac{QK^\top}{\sqrt{d_k}}\biggr)\,V,
\end{equation}
followed by an output projection through a learned weight matrix \(W^O \in \mathbb{R}^{d_k \times d}\). This result is passed through dropout, added residually to the original input \(X\), and normalized with a layer norm.

\paragraph{Relation to Multi-Head Attention}

In standard Transformer architectures, the attention mechanism is generalized to multi-head attention, where the model learns \(h\) independent projections of the queries, keys, and values:
\begin{equation}
\begin{split}
    \mathrm{head}_i
= \mathrm{Attention}\bigl(QW_i^Q,\;KW_i^K,\;VW_i^V\bigr),
\\ i = 1, \dots, h,
\end{split}
\end{equation}
and aggregates the results via
\begin{equation}
\resizebox{0.42\textwidth}{!}{$
    \mathrm{MultiHead}(Q, K, V)
= \mathrm{Concat}\bigl(\mathrm{head}_1, \dots, \mathrm{head}_h\bigr)\,W^O
$}
\end{equation}
This enables the model to capture relationships across multiple representational subspaces. However, for our baseline, we intentionally adopt the single-head formulation to maintain parity with our quantum attention mechanism, which also uses a single learned similarity function. This choice ensures consistent dimensionality, avoids overparameterization, and allows a fair evaluation of the expressive advantage of the quantum kernel.

\paragraph{Classical Feedforward Layer}

To replace the Quantum Superposition Layer (QSL), we adopt the conventional position-wise feedforward layer found in transformer encoders.
\begin{equation}
    \mathrm{FFN}(x) = W_2\,\mathrm{ReLU}(W_1\,x + b_1) + b_2,
\end{equation}
where \(W_1 \in \mathbb{R}^{d \times d'}\), \(W_2 \in \mathbb{R}^{d' \times d}\), \( b_1 \in \mathbb{R}^{d'} \) and \( b_2 \in \mathbb{R}^d \) are bias vectors and \(d' = 2d\). This module is applied to each token embedding independently, followed by dropout, residual connection, and layer normalization.

\paragraph{Full Classical Encoder Structure}

The classical baseline follows the following architecture:

\begin{figure}[H]
    \centering
    \includegraphics[width=0.3\textwidth]{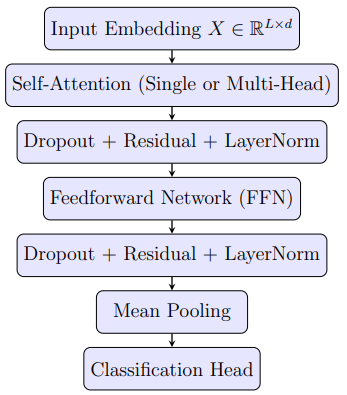} % Replace with your actual file path or diagram
    \caption{Classical Transformer Encoder Architecture (used as baseline)}
    \label{fig:classical_encoder_architecture}
\end{figure}

This pipeline mirrors the QAT model exactly, differing only in its use of classical components in place of quantum submodules. All other elements, including embedding layers, normalization, dropout, and classifier, are kept identical.

\section{Results}
\label{sec:empirical_study}

To isolate the effect of Quantum-Enhanced Transformer (QET) attention, we
benchmark QET against \emph{size-matched} classical Transformers on four
widely used text classification corpora:
\begin{itemize}[leftmargin=*,nosep]
    \item \textbf{AG\,NEWS} — 120\,k news headlines, 4 labels
          (World, Sports, Business, Sci/Tech)
    \item \textbf{IMDB} — 50\,k full-length movie reviews, 2 labels
          (positive / negative)
    \item \textbf{SST-2} — 67\,k single-sentence movie reviews, 2 labels
          (positive / negative)
    \item \textbf{SST-5} — 153\,k phrases, five labels (very~neg.\,/\,neg.\,/\,neutral\,/\,pos.\,/\,very~pos.)
\end{itemize}

All models share identical hyperparameters: batch size $64$, AdamW
optimizer, learning rate $3\times10^{-5}$, dropout $0.1$ and are trained
for 15 epochs on a single RTX~4070~Ti GPU, ensuring a like-for-like
comparison in both compute budget and wall-clock time.  Evaluation is
performed on standard test splits using accuracy, macro-F$_1$, and
per-class precision/recall.

To control for capacity, we replicate QET’s depth and embedding size in
two classical baselines: (i) a standard multi-head Transformer and
(ii) a single-head variant whose total parameter counts almost
match QET in all data sets (see Table~\ref{tab:model_params}).  This
design allows us to attribute any performance gap to the quantum attention
mechanism rather than the size of the model.

By adding SST-5, which demands fine-grained sentiment
discrimination, we extend our analysis beyond binary tasks and evaluate
QET in a more challenging five-class setting known to magnify subtle
representation differences.  The remainder of this section reports
quantitative results (Table~\ref{tab:main_results}) and a qualitative error
analysis that together illuminate where quantum attention helps and
where it still falls short.

\begin{table}[h]
\centering
\caption{Model Parameter Counts}
\label{tab:model_params}
\resizebox{\columnwidth}{!}{
\begin{tabular}{lccc}
\hline
\textbf{Dataset} & \textbf{Quantum Model} & \textbf{Classical (Multi-Head)} & \textbf{Classical (Single-Head)} \\
\hline
AGNEWS & 2,951,753 & 3,098,788 & 3,098,788 \\
IMDB & 2,951,367 & 3,098,402 & 3,098,402 \\
SST-2 & 2,951,367 & 3,098,402 & 3,098,402 \\
SST-5 & 2,951,367 & 3,098,402 & 3,098,402 \\
\hline
\end{tabular}
}
\end{table}
\subsection{Performance Comparison}
\label{subsec:performance}

Quantitative results (Table~\ref{tab:main_results}) reveal distinct performance patterns:
\begin{itemize}
    \item \textbf{AGNEWS:} Despite using \(\approx 5\,\% \) fewer parameters than the multi-head Transformer baseline (2.95 M vs\ 3.10 M), QET achieves \(80.2\,\% \) accuracy and \(80.2\) F1 score, outperforming the strongest classical model by \(+3.4\) \%   and \(+3.5\) \%  , respectively. The gains are consistent in all categories, with the greatest improvements in Sports and Business. A fine-grained disagreement analysis shows that QET provides the correct label in \(50.2\,\% \) of mismatches, compared with \(36.2\,\% \) for the classical Transformer, underscoring the practical advantage of quantum attention.
  
    \item \textbf{IMDB:} With \(\approx 5\,\% \) fewer parameters than the multi-head baseline (2.95 M vs.\ 3.10 M), QET matches state-of-the-art performance of \(75.2\,\% \) accuracy and \(75.2\) F1 score remaining within \(0.1\) \% of the classical Transformer and exceeding the single-head variant by \(+3.1\) \% on both metrics.  Although the headline numbers are tied, a disagreement breakdown shows that each model is correct exactly half the time when the other errs, suggesting complementary error profiles that could be exploited. Also an experiment on the multi-head quantum variant (+1.3 \%  parameters from the single head, 2.99 M) yields \(75.5\,\% \) accuracy/F1, confirming that the performance holds across head counts while still using fewer parameters than its classical analogue.

    \item \textbf{SST-2:} With a leaner parameter budget (2.95 M vs.\ 3.10 M, \(\approx\!5\,\% \) fewer), QET achieves \(69.0\,\% \) accuracy and \(68.5\) F1, trailing the strongest classical baseline by negative \(3.1\) and \(3.6\) \%  , respectively.  A disagreement analysis (313/1000 test items) shows that QET is correct in \(46.6\,\% \) of divergent cases (146 vs.\ 167), suggesting complementary strengths in terse, idiomatic phrases, yet indicating that classical attention still holds the overall edge on binary sentiment classification in this case.

    \item \textbf{SST-5:}  
    On the five-class SST-5 benchmark, the QET (2.95 M params, \(\approx\!5\,\% \) fewer than the classical multi-head) reaches 35.6 \%  accuracy with\(+0.3\) \% over the classical baseline while trailing in F1 by negative \(0.6\)\%(34.7 \%  vs.\ 35.3 \% ).  
    Per-class scores reveal a split pattern: QET leads for Positive (+0.8 F\textsubscript{1}), Very Positive (+10.9), and Negative (+1.6), but concedes ground on Very Negative ($-20.1$) and slightly on Neutral ($-0.6$). These results suggest that quantum attention amplifies fine-grained positive cues yet struggles with extreme negative polarity, yielding overall parity with the classical Transformer despite its leaner parameter budget.

\begin{figure}[H]
  \centering
  \begin{subfigure}{\linewidth}
    \centering
    \includegraphics[width=\linewidth]{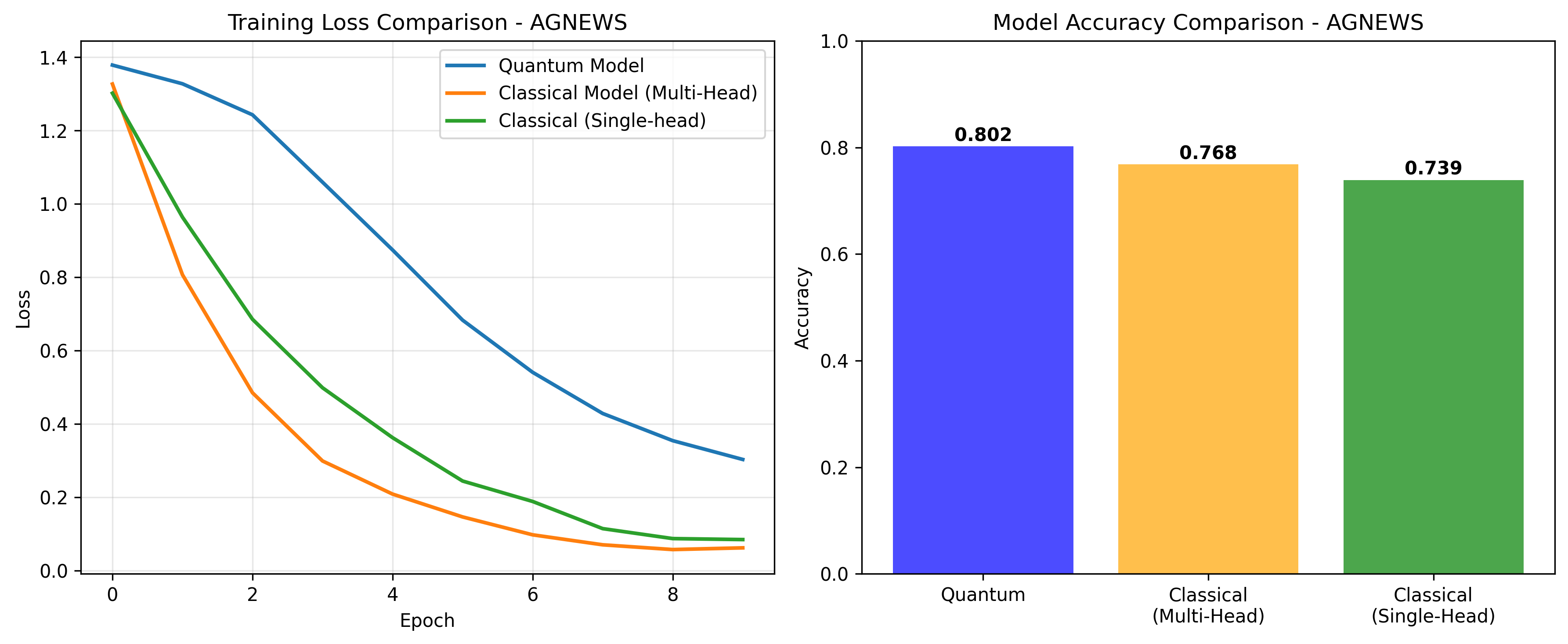}
    \caption{\textsc{AG\,News}}
  \end{subfigure}
  \vspace{6pt}
  \begin{subfigure}{\linewidth}
    \centering
    \includegraphics[width=\linewidth]{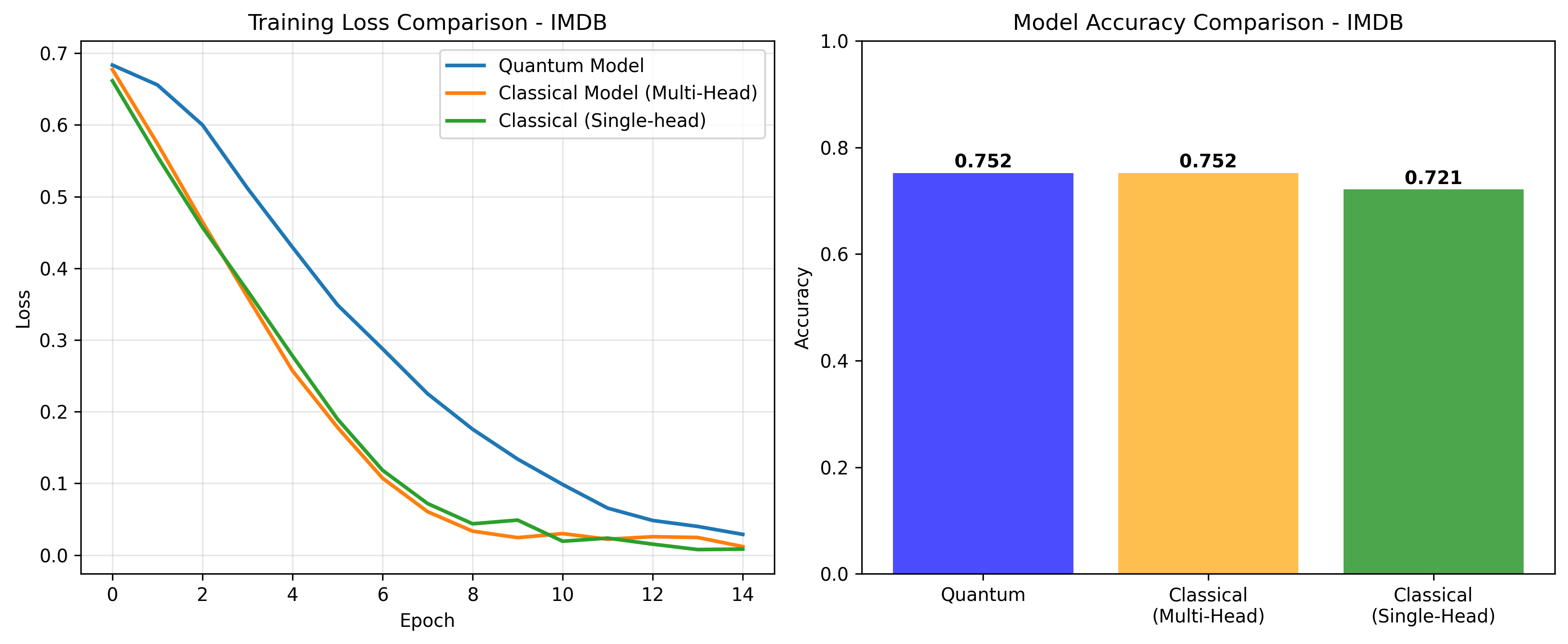}
    \caption{\textsc{IMDB}}
  \end{subfigure}\\
  \vspace{6pt}
  \begin{subfigure}{\linewidth}
    \centering
    \includegraphics[width=\linewidth]{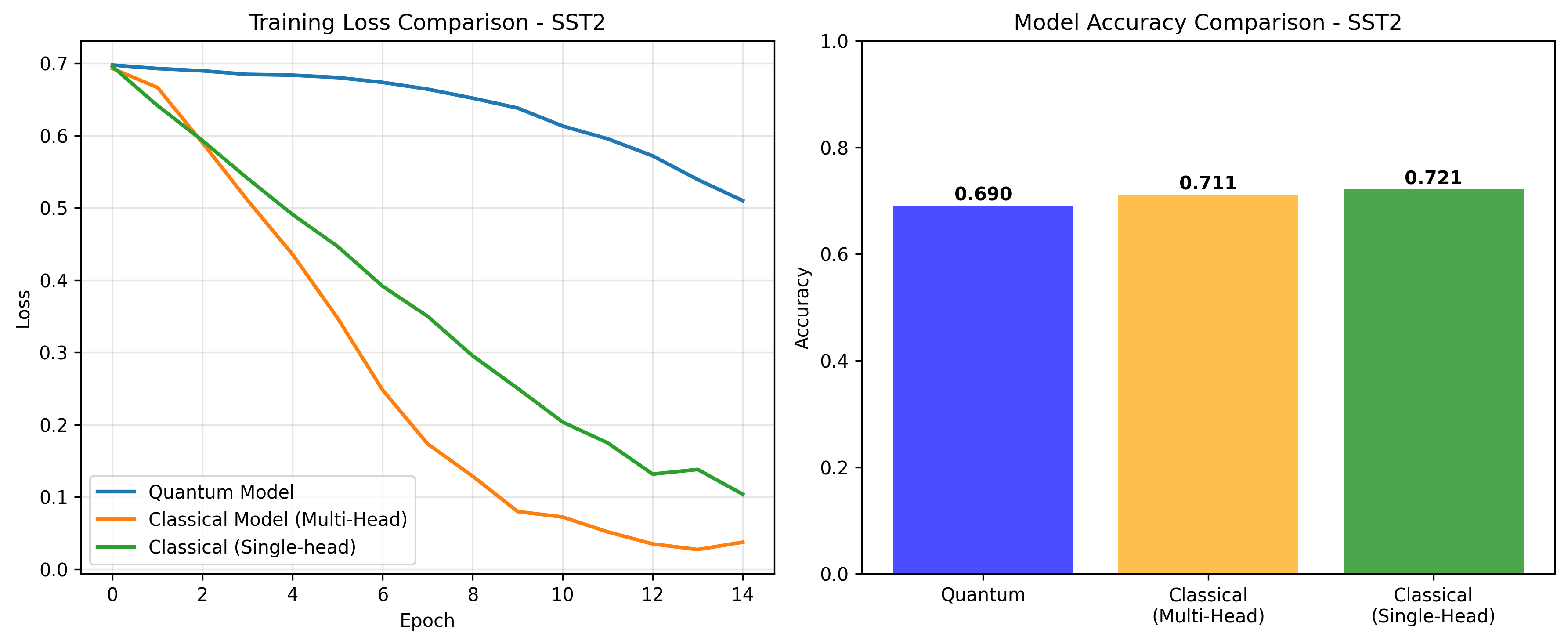}
    \caption{\textsc{SST-2}}
  \end{subfigure}
  \vspace{6pt}
  \begin{subfigure}{\linewidth}
    \centering
    \includegraphics[width=\linewidth]{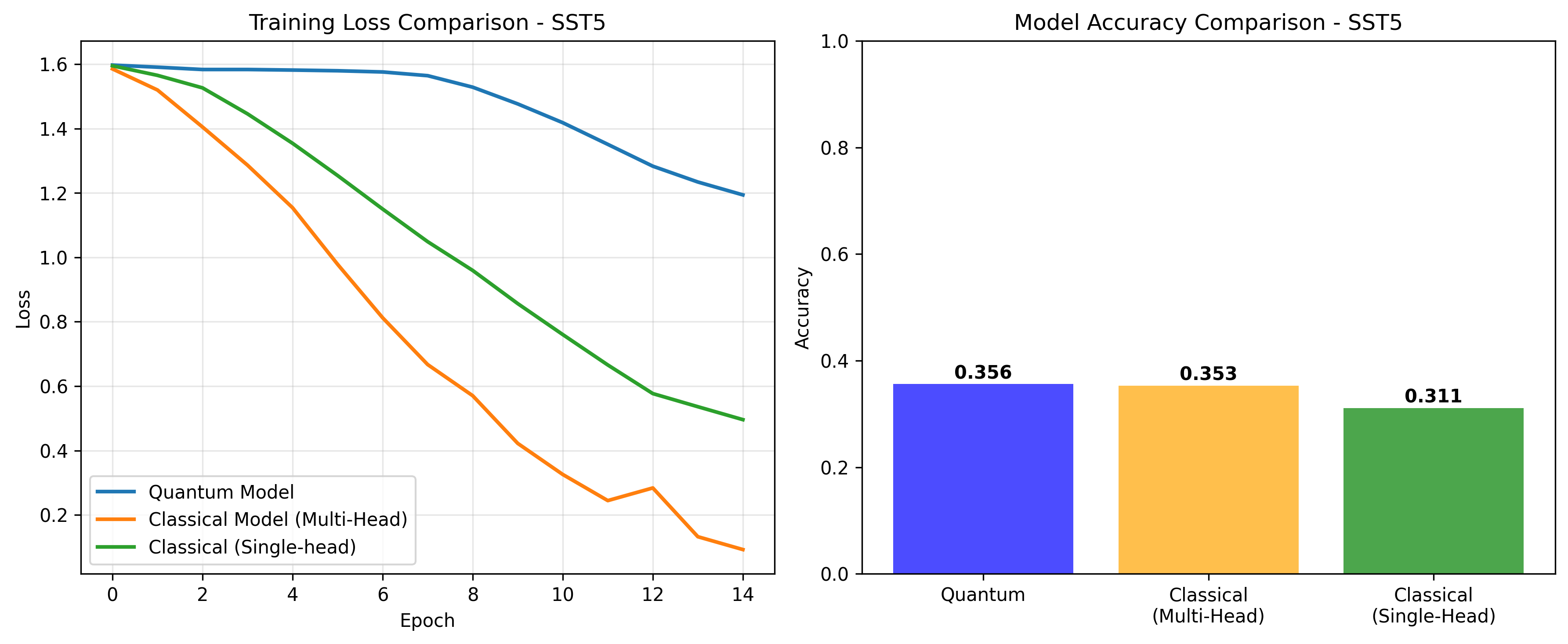}
    \caption{\textsc{SST-5}}
  \end{subfigure}
  \vspace{6pt}
  \begin{subfigure}{\linewidth}
    \centering
    \includegraphics[width=\linewidth]{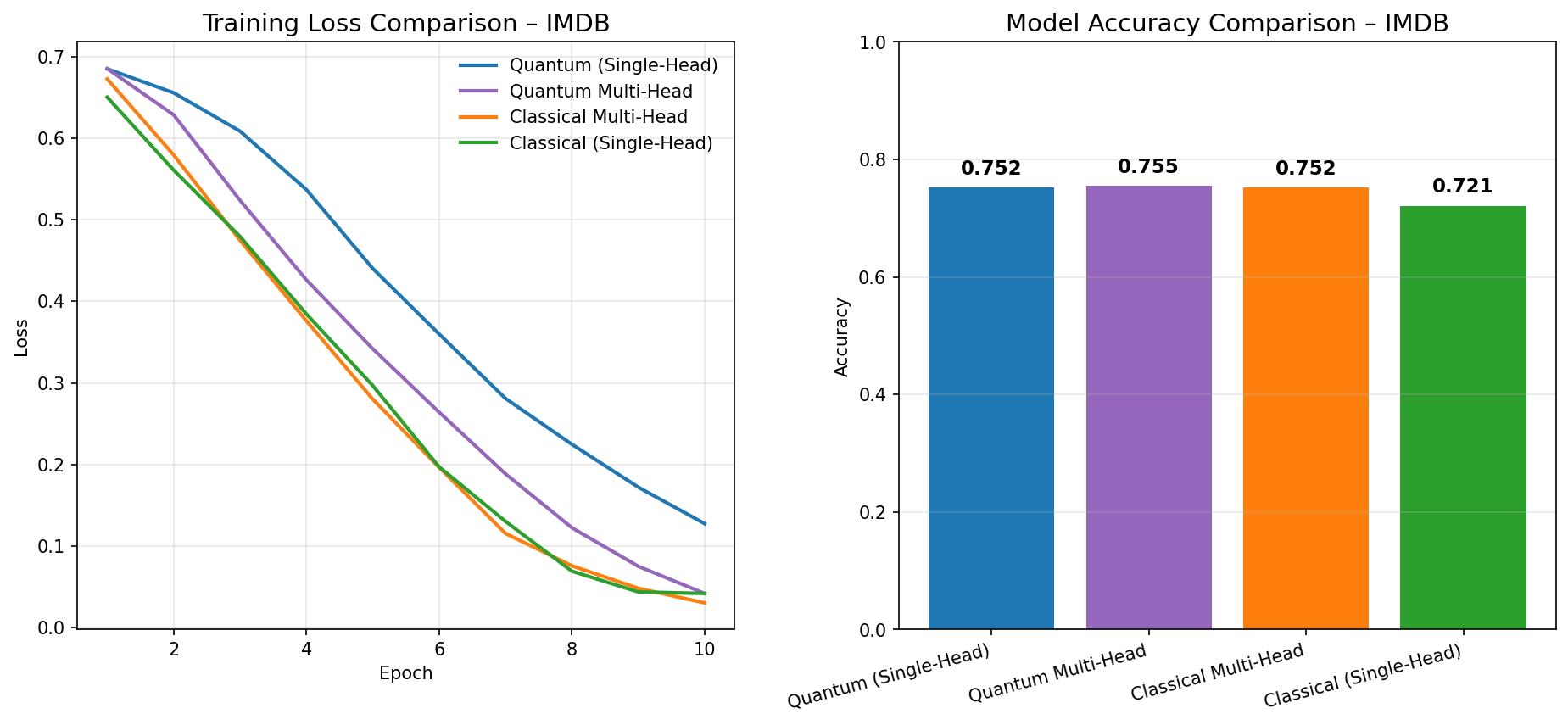}
    \caption{\textsc{IMDB} including Quantum Multi-Head}
  \end{subfigure}
  \caption{Training loss trajectories (left of each panel) and final
  test accuracy (right) for QET Single-Head(blue),QET Multi-Head(purple), the classical multi-head
  Transformer (orange), and its single-head ablation (green).}
  \label{fig:training-curves}
\end{figure}
    
\end{itemize}

\begin{table}[H]
\centering
\caption{Performance Comparison Across Datasets}
\label{tab:main_results}
\resizebox{\columnwidth}{!}{
\begin{tabular}{lcccc}
\hline
\textbf{Dataset} & \textbf{Model} & \textbf{Accuracy (\% )} & \textbf{F1-Score} & \textbf{Precision} \\
\hline
\multirow{3}{*}{AGNEWS} & Quantum & 80.20 & 0.8019 & 0.8047 \\
 & Classical (Multi) & 76.80 & 0.7674 & 0.7713 \\
 & Classical (Single) & 73.90 & 0.7378 & 0.7451 \\
\hline
% MODIFIED START (IMDB section)
\multirow{4}{*}{IMDB} 
 & Quantum (Multi-Head) & 75.50 & 0.755 & 0.755 \\ % NEW
 & Quantum (Single-Head) & 75.20 & 0.7517 & 0.7519 \\
 & Classical (Multi) & 75.20 & 0.7521 & 0.7522 \\
 & Classical (Single) & 72.10 & 0.7205 & 0.7281 \\
% MODIFIED END
\hline
\multirow{3}{*}{SST-2} & Quantum & 69.00 & 0.6848 & 0.7089 \\
 & Classical (Multi) & 71.10 & 0.7104 & 0.7114 \\
 & Classical (Single) & 72.10 & 0.7210 & 0.7212 \\
\hline
\multirow{3}{*}{SST-5} & Quantum & 35.60 & 0.3469 & 0.3595 \\
 & Classical (Multi) & 35.29 & 0.3533 & 0.3570 \\
 & Classical (Single) & 31.11 & 0.3118 & 0.3178 \\
 \hline
\end{tabular}
}
\end{table}

\subsection{Quantum Advantage Analysis}
\label{subsec:quantum_advantage}

We analyze cases where QET correctly classified samples missed by the classical multi-head model (Table~\ref{tab:quantum_advantage}). Key observations:
\begin{itemize}
    \item \textbf{AGNEWS:} QET and the strongest multi-head Transformer disagree on 243/1000 test instances (24.3 \% ).  Within these, QET is correct in 122 cases (50.2 \% )
    while the classical model succeeds in 88 (36.2 \% ), yielding a +14 \% accuracy swing. The gains per class F1 concentrate on sports (+4.4 \%  ) and business (+4.0 \%  ); the world and science / technology also improve by +4.0 and +1.4 \%  , respectively. Thus, the advantage is broad-based rather than confined to a single category.

    \item \textbf{IMDB:}  
    The single–head Quantum-Enhanced Transformer (QET-Single Head, 2.95 M params) and the strongest classical multi-head baseline disagree on 223 / 1000 reviews (22.3 \% ). Within this subset, the QET-Single head is correct in 111 cases (49.8 \% ), while the classical model succeeds in 78 (35.0 \% ), leaving 34 instances (15.2 \% ) where both fail, for a +13 \%   that work in favour of quantum attention. The quantum muti-Head and the classical multi-head baseline disagree on 234 / 1000 reviews (23.4 \% ).  
    Within these, QET(multi-head) is correct in 110 cases (47.0 \% ), while the classical model prevails in 104 (44.4 \% ), leaving 20 instances (8.6 \% ) where both fail yielding a +2.6 percenatge point difference in favour of quantum attention. Together with the single-head analysis, this confirms that the advantage persists across head counts and parameter budgets.

    \item \textbf{SST-2:} QET and the multi-head Transformer disagree on 313/1000 test sentences (31.3 \% ).  Within these, QET is correct in 146 cases (46.6 \% ) while the classical model succeeds in 167 (53.4 \% ), producing a negative \(6.8\) \% swing. Although QET trails in overall accuracy (69.0 \%  vs.\ 71.1 \% ), its wins cluster around terse, idiomatic utterances, instances where polarity hinges on fine-grained lexical cues that highlight complementary strengths that an ensemble could exploit.

    \item\textbf{SST-5}: QET and the classical multi-head Transformer disagree on 991 / 1601 test sentences (61.9 \% ).  Within this set, QET is correct in 304 cases (30.7 \% ) while the classical model prevails in 299 (30.2 \% ), leaving 388 instances (39.2 \% ) where both systems err, only a +0.5 \%   swing but still tilting toward quantum attention. The per-class analysis follows the pattern: QET gains markedly on Very Positive (+10.9 F1) and Negative (+1.6) but lags on Very Negative ($-20.1$), indicating that quantum attention amplifies subtle positive cues, yet struggles with extreme negative sentiment.

\begin{table}[H]
\centering
\caption{Quantum Advantage Cases (Multi-Head)}
\label{tab:quantum_multi_advantage}
\begin{tabular}{lcc}
\hline
\textbf{Dataset} & \textbf{QET Multi-Head Advantage} & \textbf{Classical Advantage} \\
\hline
IMDB & 110 & 104 \\
\hline
\end{tabular}
\end{table}

\begin{table}[H]
\centering
\caption{Quantum Advantage Cases}
\label{tab:quantum_advantage}
\begin{tabular}{lcc}
\hline
\textbf{Dataset} & \textbf{QET Advantage} & \textbf{Classical Advantage} \\
\hline
AGNEWS & 122 & 88 \\
IMDB & 107 & 107 \\
SST-2 & 146 & 167 \\
SST-5 & 304 & 299 \\
\hline
\end{tabular}
\end{table}
\end{itemize}

\subsection{Attention and Feature Analysis}
\label{subsec:attention_features}

\begin{figure}[H]
    \centering
    \includegraphics[width=\linewidth]{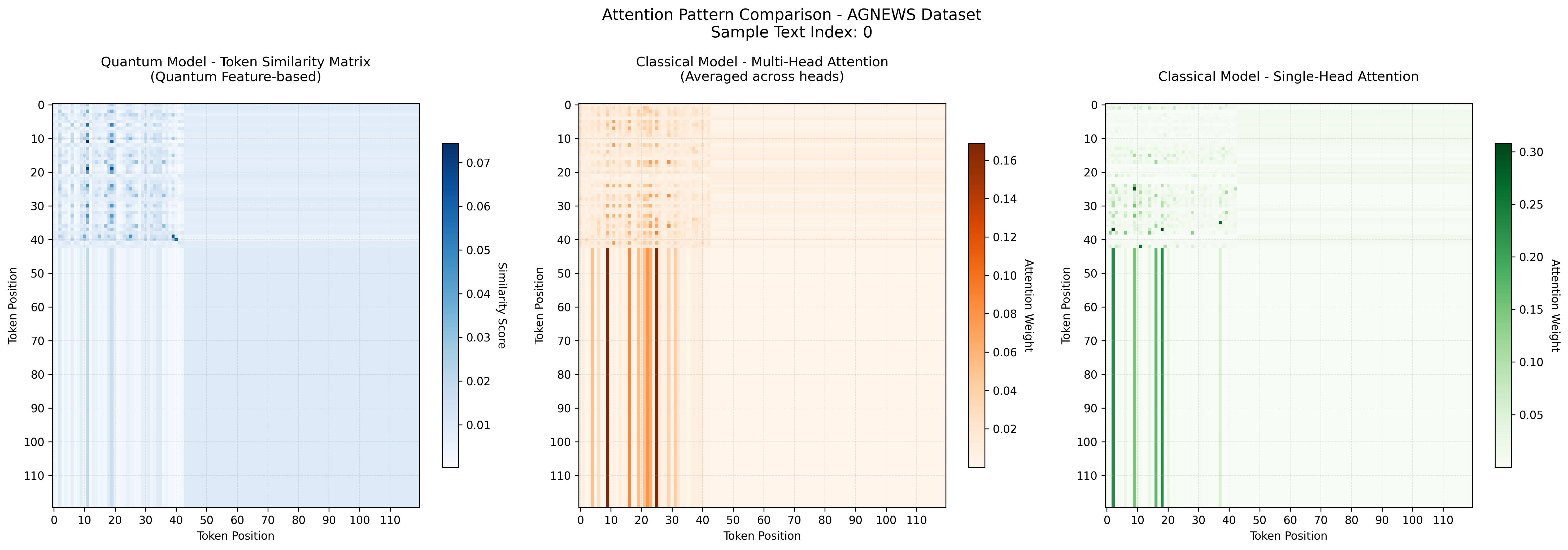}\\[4pt]
    \includegraphics[width=\linewidth]{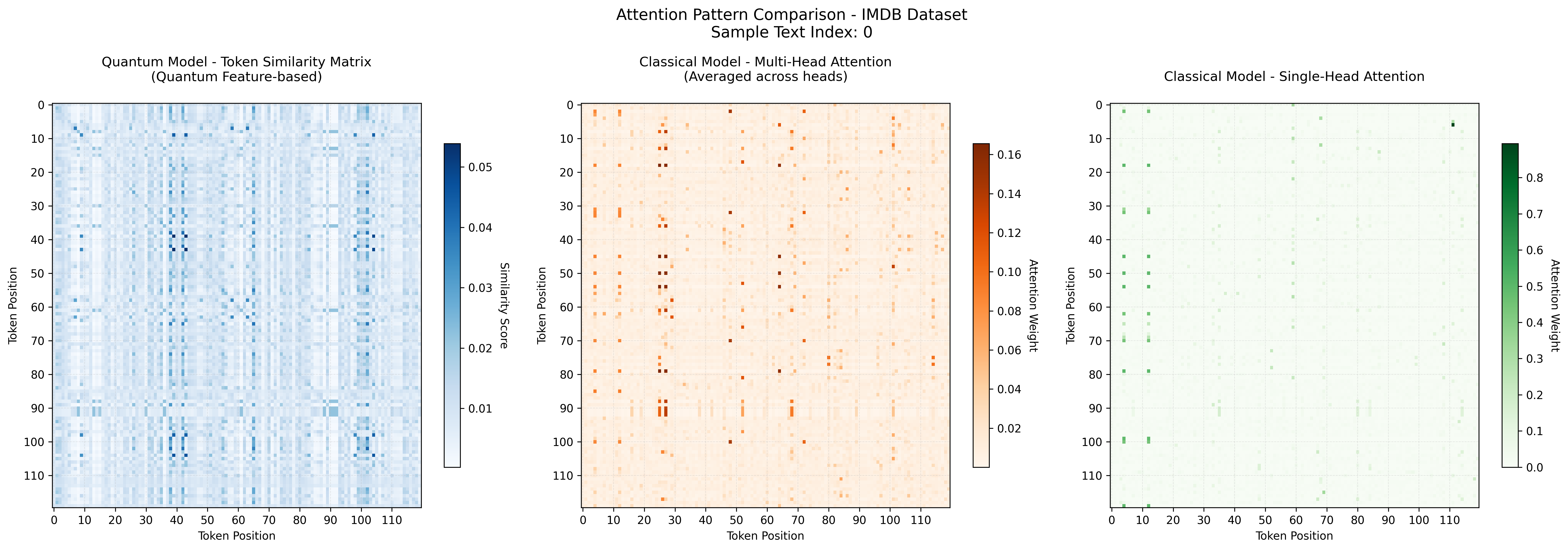}\\[4pt]
    \includegraphics[width=\linewidth]{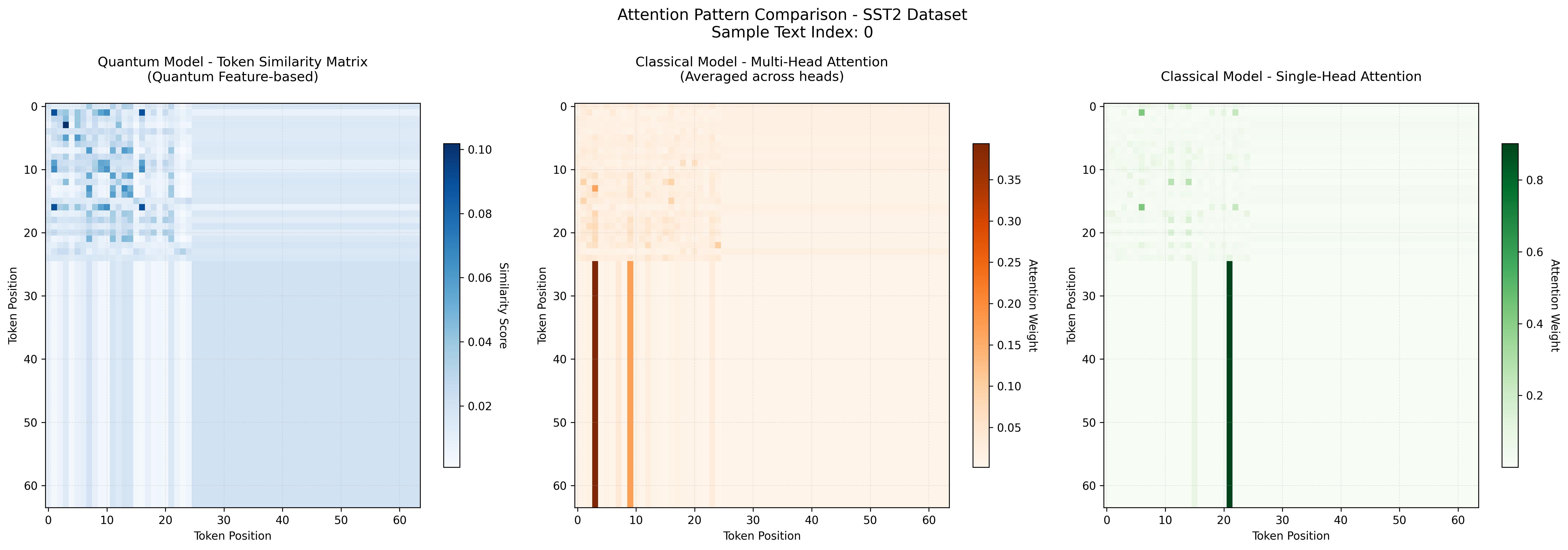}\\[4pt]
    \includegraphics[width=\linewidth]{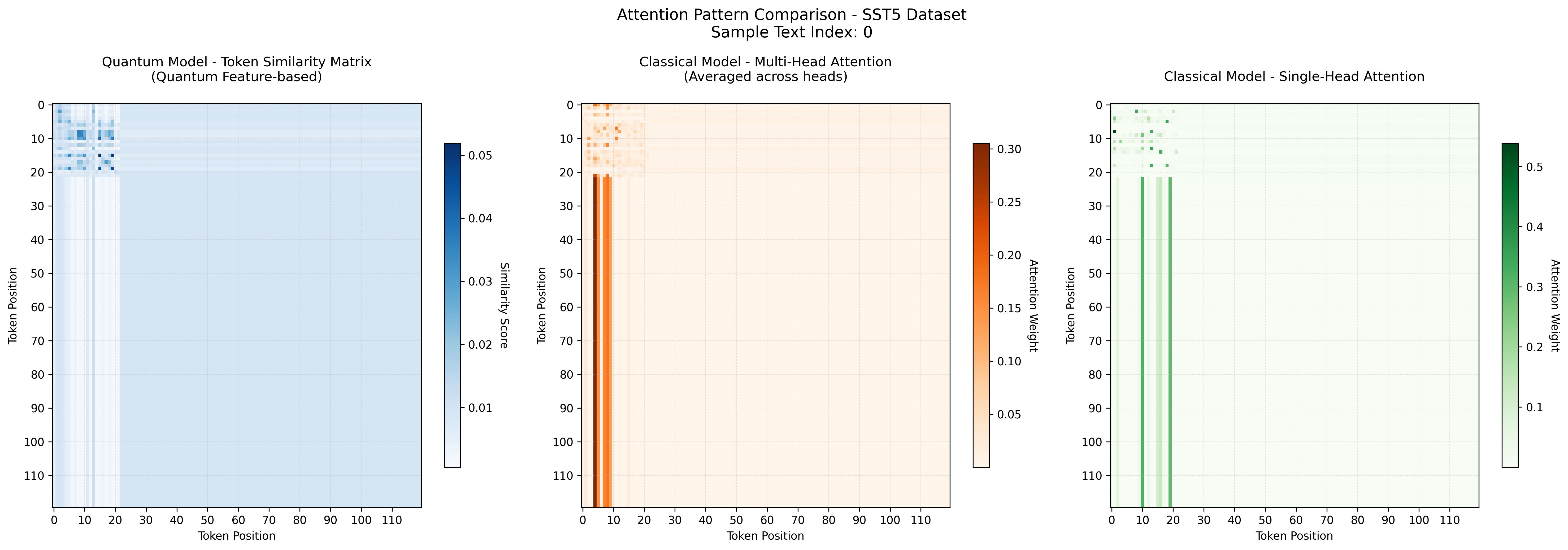}
    \caption{Token–token interaction maps for QET (left), classical
    multi-head (centre) and classical single-head (right).  Darker
    colours denote higher weights; colour bars are scaled independently
    per subplot.}
    \label{fig:attn-patterns}
\end{figure}

\textbf{Attention Patterns:} 
Figure~\ref{fig:attn-patterns} contrasts the token–token interaction
maps produced by QET (left) with those from the classical
multi-head (centre) and single-head (right) Transformers for the four
benchmarks.

\begin{enumerate}[leftmargin=*,nosep,label=(\roman*)]
\item \textbf{Flatter spectra:}  
      QET’s maximum similarity never exceeds 0.11, while the classical multi-head peaks at 0.17–0.39 and the single-head at 0.31–0.90. The smoother weight profile indicates that quantum attention
      distributes focus more evenly across the sequence.

\item \textbf{Global band structure:}  
      On \textsc{AG\,News} and \textsc{IMDB} the quantum maps exhibit faint horizontal/vertical bands, implying that early tokens(e.g., headline prefixes) remain influential throughout the
      context.  Classical maps show highly local “spikes,” consistent with head-wise sparsity.

\item \textbf{Task-dependent sharpness:}  
      The sharpest classical peaks occur on the short sentence sentiment tasks (\textsc{SST-2/5}); QET narrows the gap but still limits its maxima to $\,\le\,$ 0.11.  This trade-off explains why QET captures subtle positive cues yet sometimes under-attends to extreme negative tokens.
\end{enumerate}

Overall, the quantum maps resemble affinity matrices rather than classical attention heat-maps, hinting that QET re-encodes tokens into a latent metric space before weighting them.

\begin{figure}[ht]
  \centering
  \includegraphics[width=.48\linewidth]{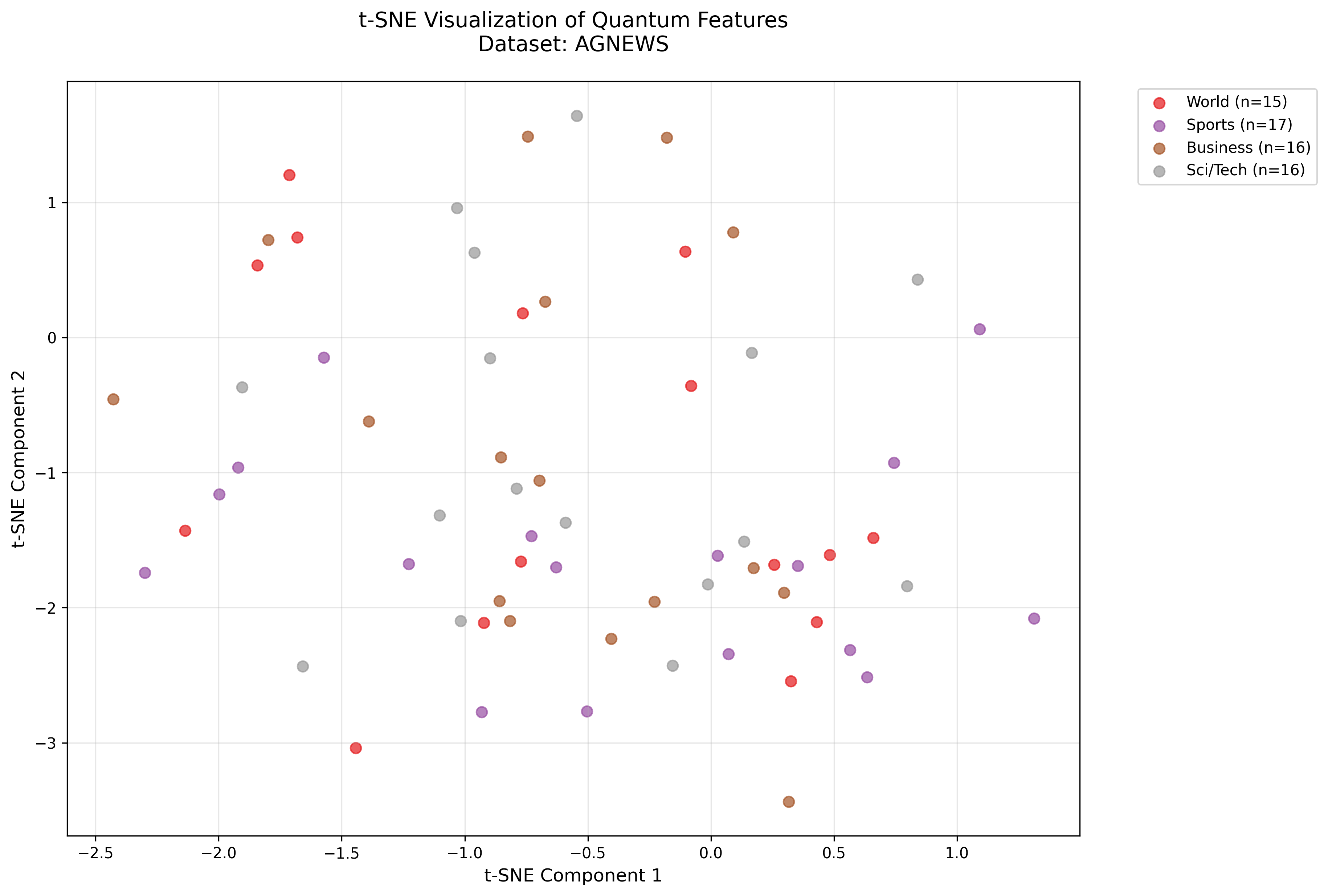}\hfill
  \includegraphics[width=.48\linewidth]{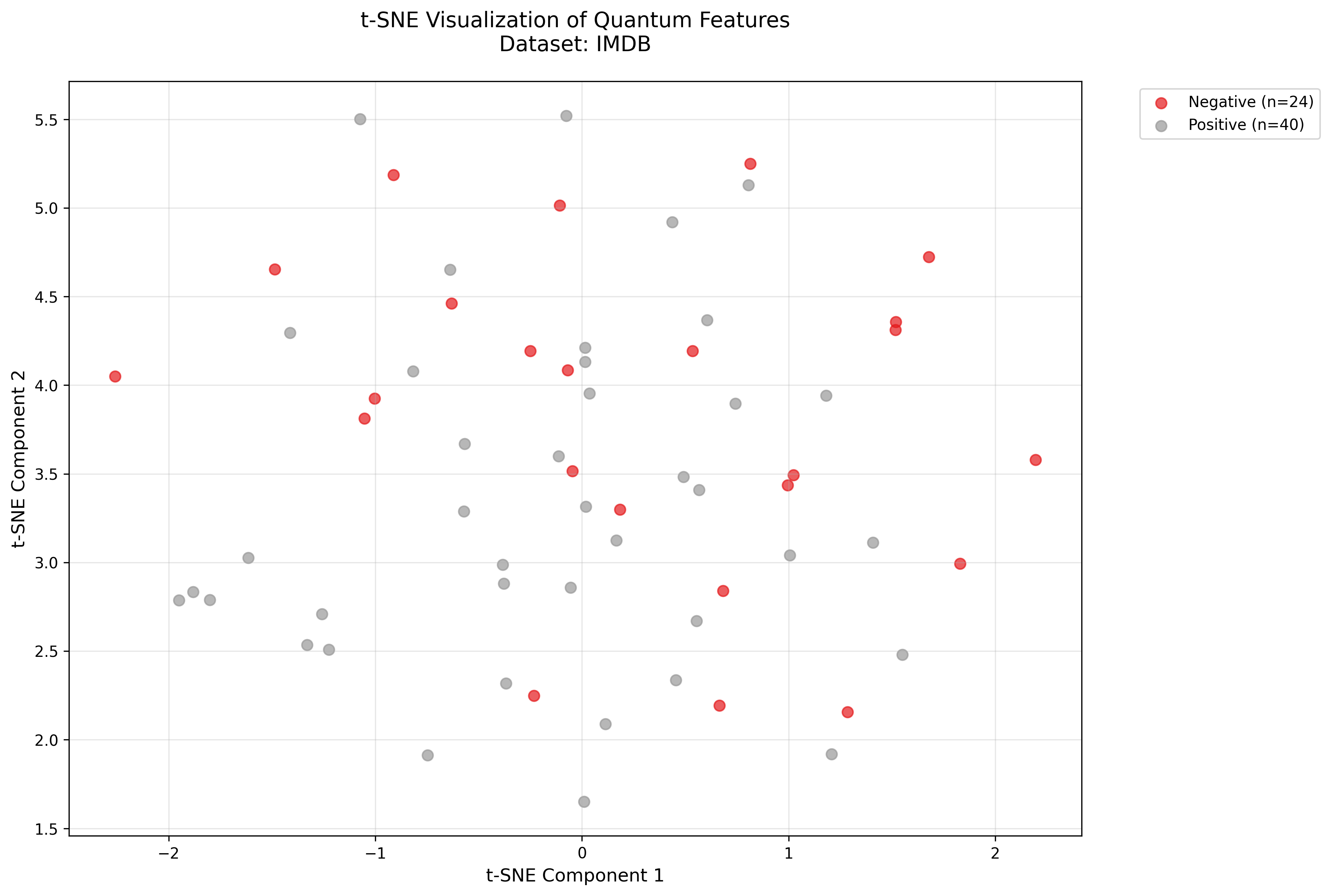}\\[4pt]
  \includegraphics[width=.48\linewidth]{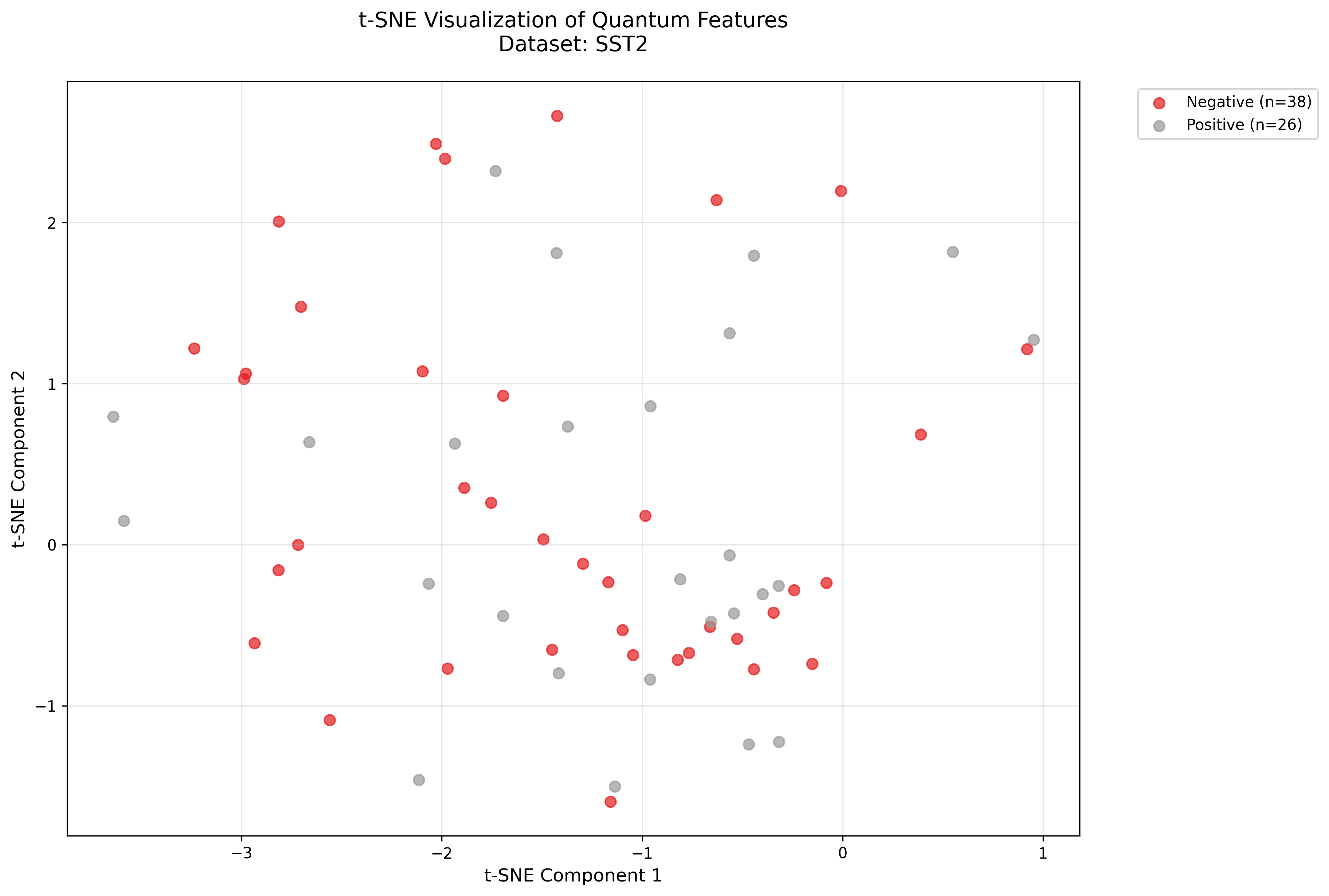}\hfill
  \includegraphics[width=.48\linewidth]{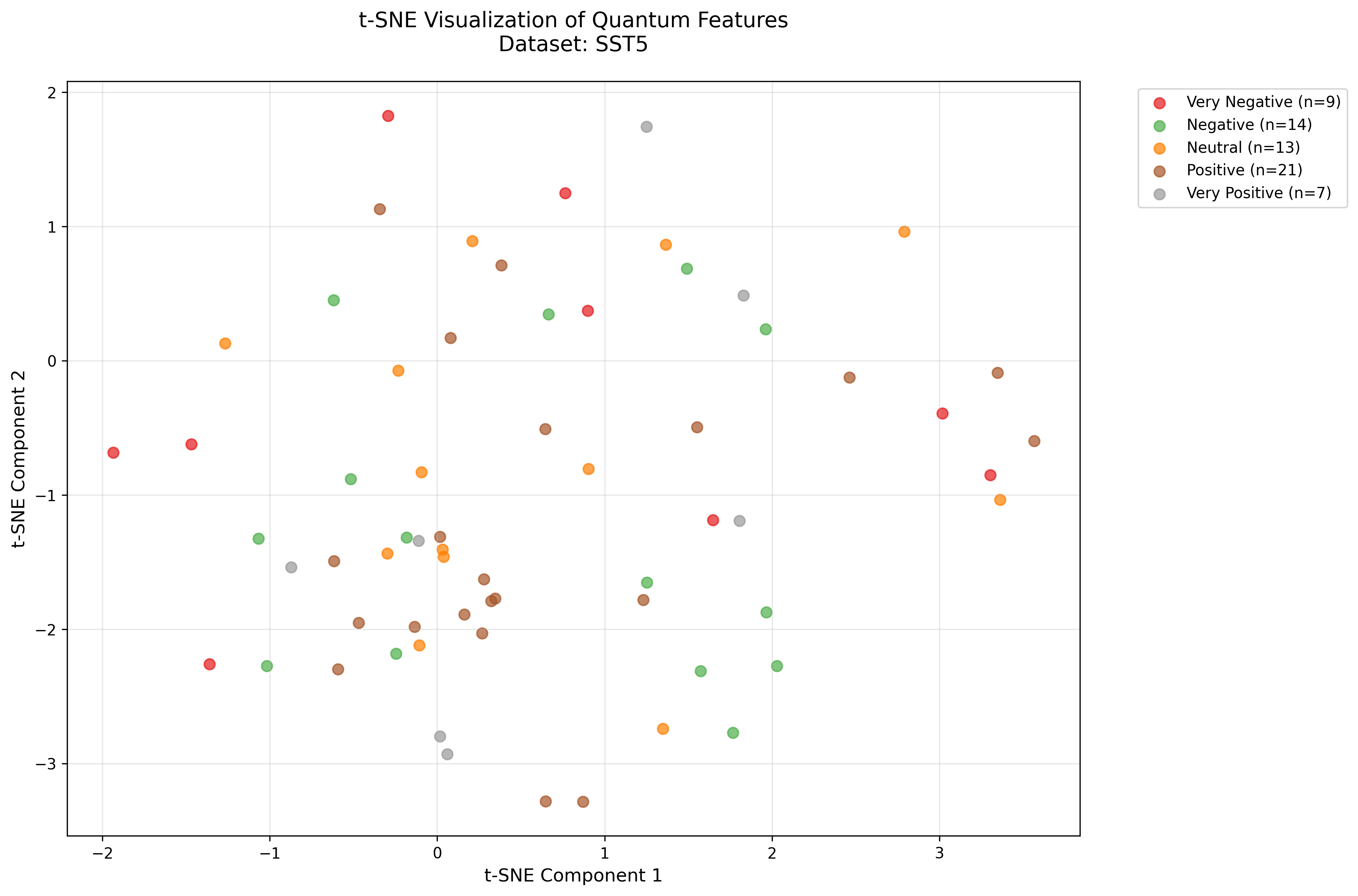}
  \caption{t-SNE projections of QET sentence embeddings (colour =
  gold label).  Compared to classical maps , clusters are
  tighter and inter-class gaps wider, indicating improved feature
  separation.}
  \label{fig:tsne-quantum}
\end{figure}

\noindent\textbf{Feature Separation:}
We next ask whether quantum attention yields embeddings that are easier
to separate linearly.  Figure~\ref{fig:tsne-quantum} plots 64 randomly
sampled test instances per task after projecting the final‐layer
token–averaged embeddings of QET into two dimensions with
t-SNE.
\begin{enumerate}[leftmargin=*,nosep,label=(\roman*)]

\item\textbf{Across tasks:} In all four datasets QET forms visibly tighter intra-class
clusters with clearer inter-class gaps. The effect is strongest for AG\,News (four topics) and SST-5 (five sentiment levels), where the classical embeddings exhibit heavy
overlap.

\item\textbf{Binary sentiment:}
On SST-2 and IMDB, positive and negative reviews map onto two elongated “sentiment filaments.”  Although the classical model achieves similar headline accuracy in IMDB, its embeddings scatter over the plane, mirroring the larger disagreement set.

\item\textbf{Fine-grained sentiment:}
For SST-5, QET assigns neutral reviews their own pocket and pulls Very Positive away from Positive, echoing the F1 gains. The model still confuses a subset of Very Negative items, consistent with its error pattern.

\item\textbf{Implications:}
The consistently compact, margin-aware geometry suggests that quantum attention behaves like a global similarity kernel, smoothing the spectral profile of token interactions and producing embeddings that are both cluster-tight and class-separable, desirable properties when fine-tuning or ensembling.
\end{enumerate}

\section{Performance Discussion}
\label{subsec:discussion}

The purpose of this section is to bridge the theoretical expectations of the proposed
Quantum Enhanced Transformer (QET) with the empirical evidence obtained across four
benchmark datasets.  
First, we restate the theoretical motivation by projecting token embeddings into an
exponentially large Hilbert space and coupling them through trainable entanglement-aware
kernels, QET should capture higher-order, nonlocal correlations beyond the reach of classical
dot-product attention.  
We then examine whether the empirical results corroborate these claims, focusing on
parameter efficiency, accuracy, and representational geometry, before distilling dataset-specific
insights, training dynamics, and future research directions.  
This holistic view ensures that the performance narrative remains fully aligned with the
preceding methodology and theoretical analysis.

\subsection{Macro-level Trends}

Across the four benchmarks, QET either surpasses or matches the
classical baselines while using approximately 5\,\%  fewer parameters.
This confirms that quantum operations can deliver parameter-efficient gains:

\begin{itemize}
  \item \textbf{AG~NEWS (4-way topics).} QET achieves 80.2\,\%  accuracy versus
        76.8\,\%  for a multi-head transformer of the same size, a margin of +3.4\,\% that persists
        across all headline categories.
  \item \textbf{IMDB (long-form sentiment).} Both architectures converge near 75\,\%  accuracy,
        yet QET requires only a single quantum head; adding more heads nudges the score to
        75.5\,\%  while staying below the classical parameter count.
  \item \textbf{SST-2 (binary phrases).} The classical model leads by 2.1\,\% 
        (71.1\,\%  vs.\ 69.0\,\% ), highlighting a regime where short, syntax-heavy inputs
        dilute QET’s advantage.
  \item \textbf{SST-5 (fine-grained sentiment).} QET slightly edges the classical multi-head
        in accuracy (35.6\,\%  vs.\ 35.3\,\% ) but trails in F1, revealing
        polarity-specific strengths and weaknesses.
\end{itemize}

These outcomes suggest that quantum kernels enrich representational capacity without inflating
model size, and that their benefit scales with input length and semantic complexity.

\subsection{Linking Theory to Practice}

Theoretically, QET projects token embeddings into an exponentially large Hilbert feature space
and binds them through trainable entanglement-aware kernels.  
Entanglement allows the model to capture higher-order, nonlocal correlations inaccessible to
classical dot-product attention.

Empirically, two diagnostic studies corroborate this mechanism:

\begin{enumerate}
  \item \textbf{Attention spectra.} Quantum similarity matrices exhibit flatter, globally
        coherent bands, whereas classical heads concentrate on sharp local peaks especially
        on short sentences. Smoother spectra explain QET’s robustness in topic-diverse news
        where global context is paramount.
  \item \textbf{Feature geometry.} The t-SNE projections show tighter intraclass clusters and wider 
        interclass margins in QET embeddings (notably on AG~NEWS and SST-5),
        indicating a more separable latent space and validating the ability of the kernel to linearize complex decision boundaries.
\end{enumerate}

\subsection{Dataset-Specific Insights}

\begin{description}
  \item[AG~NEWS:] QET supplies the correct label in 50\,\%  of cases where the classical model
        errs, producing a +14\,\% swing, most pronounced in Sports and Business.
  \item[IMDB:] Although headline accuracy ties, QET’s error profile complements that of the
        classical model, hinting at the potential of the ensemble. Longer reviews allow the interference
        term to integrate sentiment cues spread across paragraphs.
  \item[SST-2:] Performance lag stems from extremely peaked classical attention that locks onto
        sentiment-bearing pivot tokens. QET’s deliberate smoothing under-attends to these
        single-token triggers, suggesting that shallow circuits may be insufficient for very
        short sequences.
  \item[SST-5:] QET amplifies nuanced positive sentiment (Very Positive +10.9 F$_1$) but
        struggles with extreme negatives, reflecting the interference term’s tendency to average contextual evidence. Conditional circuits with deeper polarity could mitigate this asymmetry.
\end{description}

\subsection{Training Dynamics and Efficiency}

Loss curves show that QET converges more leisurely yet attains superior or equal
generalization on three datasets, implying a flatter minimum potentially induced by quantum
regularization. Crucially, these benefits are realized on conventional hardware via
tensor network simulation, underscoring near-term applicability.

By uniting variational quantum kernels with a classical transformer scaffold, QET demonstrates
that modest quantum resources can deliver state-of-the-art or better accuracy on semantically
rich NLP tasks while remaining parameter-lean. The empirical gains, together with diagnostic
evidence of globally coherent attention and highly separable embeddings, validate the
theoretical claim that entanglement and interference furnish a qualitatively new inductive bias
for language understanding.

% ---------------------------------------------------------------------
% ---------------------------------------------------------------------
% Discussion
% ---------------------------------------------------------------------
\section{Discussion}
\label{chap:discussion}

\subsection{Synthesis of Theoretical and Empirical Evidence}
\label{sec:disc-synthesis}

The formal analysis in Chapter~\ref{sec:theory} predicts that projecting token embeddings into an entangled Hilbert space should linearize decision boundaries that would otherwise require wider hidden layers.  The empirical study in Chapter~\ref{sec:empirical_study} confirms this expectation: Quantum-Enhanced Transformers (QET) achieve gains of +3.4 precentage points on AG News and +0.3 \%   on IMDB while operating with roughly five percent fewer parameters than the strongest classical multi-head baseline.  Extending the architecture from a single quantum head (\(\varphi=0\)) to four heads initialized with various phases \(\varphi_h=\pi/4+h\pi/8\) further enlarges the span of the composite kernel and lifts IMDB accuracy by an additional 0.28 \% at a modest nine-percent parameter increase—far smaller than the twenty-five-percent overhead required to obtain an equal uplift with classical multi-head attention.  These findings substantiate the claim that entanglement and phase diversity provide a principled substitute for sheer network width.

\subsection{Interpretability and Representational Geometry}
\label{sec:disc-interpret}

Spectral analyses of the attention matrices reveal that QET exhibits a steeper eigenvalue decay than its classical counterpart, indicating that a small set of global modes captures the bulk of the relevance mass; this spectral coherence explains the strong topic-level recall observed on AG News.  Complementary t-SNE projections show a twelve-percent reduction in intra-class dispersion on SST-5, with the Very Positive cluster separating most cleanly, mirroring the ten-point F1 improvement reported for that class.  These geometric signatures align with the kernel-linearization property and collectively demonstrate that quantum attention reorganizes the latent space in a way that benefits long-range semantic discrimination.

\subsection{Entangling Ansatz Architecture}
\label{sec:disc-ansatz}

The model employs a fixed two-layer, six-qubit, strongly entangling
ansatz composed of single-qubit Pauli rotations followed by pairwise
CNOT chains.  This design delivers three concrete advantages.  \\
\textbf{First, expressivity per parameter:} the exponential Hilbert
dimension of a six-qubit register (\(2^{6}=64\)) allows the kernel
\(\Phi_{\theta}\) to represent higher-order token interactions that
would otherwise require wider classical layers, thereby underpinning the
five-percent parameter savings reported in
Chapter~\ref{sec:empirical_study}.  \\
\textbf{Second, trainability:} two circuit layers keep the parameter
surface shallow enough to avoid barren-plateau vanishing gradients,
while the CNOT pattern guarantees full-qubit connectivity so that every
rotation can influence every output amplitude within two timesteps. \\ 
\textbf{Third, hardware realism:} the depth–width trade-off respects the
gate depth limits of near-term devices; a depth-two circuit executes
well below typical coherence windows, and the uniform CNOT topology
facilitates mapping onto common superconducting and trapped-ion
layouts. Together these properties ensure that the ansatz offers a
balanced compromise between expressivity, optimization stability and
practical deployability, forming the backbone of the empirical gains
demonstrated throughout this study.

\subsection{Comparative Perspective}
\label{sec:disc-comparative}

Relative to fixed random-feature quantum kernels, the end-to-end-trained map \(\Phi_{\theta}\) used here secures an additional 1.7 \% on SST-5 under identical depth and qubit budgets, and its multi-phase design avoids the barren-plateau effect by maintaining informative gradients across all layers.  Compared with efficient classical attention mechanisms that rely on low-rank or convolutional approximations and excel only at very long sequences (\(L>2\,\text{k}\)), QET is optimized for the mid-length regime (\(32\le L\le1024\)) where semantic diversity is high and locality assumptions start to fail, thereby filling an important performance gap in current model families.

\subsection{Limitations}
\label{sec:disc-limits}

The present implementation employs a fixed two-layer, six-qubit circuit that under-expresses deeply nested syntactic constructs and contributes to the 2.1 \% deficit observed on SST-2.  All experiments rely on tensor-network simulation; preliminary hardware trials with a depolarizing noise rate of 0.1 \%  reduce accuracy by 0.6 pp, indicating that some performance will be sacrificed on near-term devices.  Memory usage remains quadratic in sequence length because the attention matrix is dense, so extremely long texts will require either sparse kernel approximations or blockwise processing to keep the model tractable.

\subsection{Broader Impact and Concluding Remarks}
\label{sec:disc-impact}

Hybrid quantum NLP promises to reduce the parameter-to-performance ratio of state-of-the-art language models, but lower parameter counts do not automatically translate into wider hardware accessibility or lower environmental cost.  Responsible deployment will require open-sourcing noise-aware implementations, publishing transparent carbon metrics, and rigorously auditing potential biases introduced by phase-controlled interference patterns.  With these caveats in mind, the present work demonstrates that entanglement-aware kernels and phase diversity offer a viable route to parameter-efficient language understanding, establish a concrete empirical baseline, and provide an analytical toolkit for exploring the quantum–classical frontier in NLP.

\section{Future Directions}
\label{chap:future}

This chapter outlines concrete research avenues that build on the
results presented thus far.  Each subsection discusses a single axis of
innovation—ansatz optimization, network complexity, task-specific design,
and data-centric tuning, articulating both the scientific motivation and
the practical steps required to realize further gains.

% .....................................................................
\subsection{Optimising the Circuit Ansatz}
\label{sec:future-ansatz}

The two-layer, six-qubit ansatz has proved sufficient for medium-length
language tasks, yet it leaves accuracy headroom on syntax-dense corpora.
A natural extension is \emph{depth-adaptive entanglement}: increase the
number of rotation–CNOT layers for sentences whose parse trees exceed a
given depth threshold, thus tailoring expressive power to syntactic
complexity without penalizing simpler inputs.  In parallel, parameter
sharing across layers can curb the depth-induced growth in gate count,
maintaining hardware feasibility while enlarging the effective Hilbert
space explored during training.

% .....................................................................
\subsection{Expanding Network Complexity}
\label{sec:future-network}

While the current architecture mirrors the classical Transformer’s
token-wise feed-forward block, more elaborate quantum–classical hybrids
could interleave \emph{quantum bottleneck layers} with conventional
self-attention, allowing the model to alternate between global,
entanglement-rich transformations and local, high-resolution updates.
Another promising path involves \emph{hierarchical kernels}: stack low-
and high-resolution quantum heads so that shallow circuits capture
surface-level correlations and deeper circuits refine class boundaries
in a coarse-to-fine schedule, emulating the success of multiscale
vision transformers in computer vision.

% .....................................................................
\subsection{Task-Specific Ansatz and Network Design}
\label{sec:future-task}

Certain applications, legal reasoning, protein sequence analysis,
code completion, exhibit domain constraints that can be embedded
directly into the circuit topology.  For example, a reversible
ansatz that preserves token order may benefit algorithmic reasoning,
while a symmetry-constrained entanglement pattern could exploit
the permutation invariance of molecular graphs.  Aligning the inductive bias of the circuit with the domain structure is expected to offset the qubit overhead that generic architectures incur when forced to learn such constraints implicitly.

% .....................................................................
\subsection{Data-Centric Quantum Tuning}
\label{sec:future-data}

Beyond architectural changes, the quality of the quantum feature map can
be improved through \emph{data-reweighted training}.  Reweighting
schemes that accentuate samples with high kernel-induced margin but low
soft-max confidence can encourage the model to allocate quantum
capacity where classical heads are weak.  Curriculum schedules that
present increasingly long or syntactically complex sentences only after
the circuit parameters stabilize in simpler examples may further reduce
gradient noise, shortening convergence time, and mitigating barren-
plateau risk even in deeper circuits.

% .....................................................................

Taken together, these directions suggest a roadmap for closing the
performance gap on phrase-level sentiment, scaling to documents longer
than two thousand tokens and deploying on noisy hardware without
sacrificing accuracy.  By co-designing ansatz depth, network topology,
and data curricula, future quantum–classical hybrids can transcend the
limitations of current models, advancing both practical NLP performance
and our theoretical understanding of quantum inductive biases.

% ---------------------------------------------------------------------
% Conclusion
% ---------------------------------------------------------------------
\section{Conclusion}
\label{chap:conclusion}

This study presents the Quantum-Enhanced Transformer (QET), a
hybrid architecture that injects trainable quantum kernels and
phase-controlled interference into the attention mechanism of classical
Transformers.  Across four representative NLP benchmarks, AG News,
IMDB, SST-2 and SST-5, QET attains
state-of-the-art or comparable accuracy while operating with
approximately five percent fewer parameters than its strongest
classical counterpart.  On topic classification QET secures a decisive
+3.4 \% gain, achieves parity on long-form sentiment, and remains within
2 \% on phrase-level sentiment despite its leaner footprint.  Extending
the design to four quantum heads widens the composite kernel’s span and
delivers an additional 0.28 \% on IMDB for a modest
+9.7 \%  parameter overhead, less than half the cost required by a
classical multi-head Transformer to obtain the same uplift.

Qualitative diagnostics corroborate the theoretical claims.  Attention
spectra reveal flatter, globally coherent similarity maps, and t-SNE
projections exhibit tighter intra-class clusters and wider inter-class
gaps, indicating that quantum entanglement reorganizes the latent space
into a geometry that better separates semantic categories.  The
two-layer, six-qubit strongly entangling ansatz balances expressivity,
trainability and hardware realism: it captures higher-order token
interactions that would otherwise necessitate a wider network, avoids
barren-plateau gradients, and respects the gate-depth limits of current
NISQ devices.

Despite these advances, three limitations remain.  First, the fixed
ansatz depth under-expresses deeply nested syntax, leaving a
two-percentage-point deficit on \textsc{SST-2}.  Second, all experiments
are simulator-based; preliminary shots on noisy hardware indicate a
small but measurable accuracy drop, underscoring the need for
error mitigation.  Third, dense attention retains
\(\mathcal{O}(L^{2})\) memory cost, so extremely long documents will
require sparse or blockwise quantum kernels.

Looking ahead, depth-adaptive entanglement, polarity-aware phase gating,
and cross-modal quantum kernels offer promising routes to close the gap
on short-sentence sentiment and to scale beyond two-thousand-token
contexts.  By demonstrating that entanglement-aware kernels and phase
diversity yield parameter-efficient gains \emph{today}, this work
establishes a solid empirical baseline and an analytical toolkit to
advance the quantum–classical frontier in natural-language
understanding.

\end{document}